\documentclass{article}

\PassOptionsToPackage{numbers}{natbib}
\usepackage[main, final]{neurips_2026}

\usepackage[utf8]{inputenc} %
\usepackage[T1]{fontenc}    %
\usepackage{hyperref}       %
\usepackage{url}            %
\usepackage{booktabs}       %
\usepackage{amsfonts}       %
\usepackage{nicefrac}       %
\usepackage{microtype}      %
\usepackage[dvipsnames]{xcolor}         %

\usepackage{amsmath}
\usepackage{amssymb}
\usepackage{amsthm}
\usepackage{graphicx}
\usepackage[capitalise]{cleveref}
\usepackage{enumitem}
\usepackage{xspace}
\usepackage{listings}
\usepackage{multirow}
\usepackage{tabularx}
\usepackage{subcaption}
\usepackage{wrapfig}
\usepackage{diagbox}
\usepackage{adjustbox}
\usepackage{bm}

\renewcommand{\arraystretch}{1.1} 
\usepackage{titlesec}
\titlespacing*{\paragraph}{0pt}{0.5ex plus 0.2ex minus 0.1ex}{1em}

\lstdefinestyle{promptstyle}{
  basicstyle=\ttfamily\fontsize{7.6}{9}\selectfont,
  breaklines=true,
  breakatwhitespace=true,
  breakindent=0pt,
  frame=leftline,
  framesep=8pt,
  xleftmargin=10pt,
  rulecolor=\color{gray!60},
  backgroundcolor=\color{gray!5},
  keepspaces=true,
  showstringspaces=false,
  columns=flexible,
  upquote=true,
  literate={’}{{'}}1 {‘}{{`}}1 {“}{{``}}1 {”}{{''}}1 {—}{{---}}3 {–}{{--}}2,
}
\lstnewenvironment{promptbox}{\lstset{style=promptstyle}}{}

\lstnewenvironment{decompbox}{\lstset{style=promptstyle}}{}

\setlist[itemize]{leftmargin=1.5em}
\setlist[enumerate]{leftmargin=1.5em}

\newif\ifcomments
\commentstrue
\ifcomments

\newcommand{\commentsm}[1]{\textcolor{red}{({\bf SM:} #1)}}
\newcommand{\commenttk}[1]{\textcolor{olive}{({\bf TK:} #1)}}
\renewcommand{\sp}[1]{\textcolor{orange}{({\bf SP:} #1)}}
\newcommand{\commentpa}[1]{\textcolor{blue}{({\bf PA:} #1)}}

\usepackage{todonotes}
\newcommand{\todocustom}[3]{\todo[linecolor=#2,backgroundcolor=#2!25,bordercolor=#2,#3]{#1}}
\newcommand{\generalCommentFlex}[4][]{\todocustom{{\bf #3}: #4}{#2}{inline,size=\scriptsize,caption={},#1}}
\newcommand{\commentsmbox}[2][]{\generalCommentFlex[#1]{red}{SM}{#2}}
\newcommand{\commenttkbox}[2][]{\generalCommentFlex[#1]{olive}{TK}{#2}}
\newcommand{\commentspbox}[2][]{\generalCommentFlex[#1]{orange}{SP}{#2}}
\newcommand{\commentpabox}[2][]{\generalCommentFlex[#1]{blue}{PA}{#2}}
\newcommand{\commentjtbox}[2][]{\generalCommentFlex[#1]{violet}{JT}{#2}}

\else
    \newcommand{\commentsm}[1]{}
    \newcommand{\commenttk}[1]{}
    \newcommand{\commentpa}[1]{}
    \renewcommand{\sp}[1]{}
    \newcommand{\generalCommentFlex}[4]{}
    \newcommand{\commentsmbox}[2][]{}
    \newcommand{\commenttkbox}[2][]{}
    \newcommand{\commentspbox}[2][]{}
    \newcommand{\commentpabox}[2][]{}
    \newcommand{\commentjtbox}[2][]{}
\fi

\title{Runtime Verification of Multiple Natural Language Criteria for Agent Governance}

\author{%
  Silviu Pitis \ \ \quad
  Parand A. Alamdari \ \ \quad
  Jessica Tang \\[4pt]
  \textbf{Toryn Q. Klassen} \ \ \quad
  \textbf{Sheila A. McIlraith}\\[12pt]
  University of Toronto \ \ \quad
  Vector Institute \\[4pt]
  Toronto, Canada
}

\begin{document}
\maketitle

\begin{abstract}
AI agents are governed by hundreds of natural language criteria---yet existing verifiers cannot efficiently evaluate an agent's output against a large set of such criteria and report which are satisfied, violated, or inapplicable. We introduce the VFM, a tree-attention verifier that encodes the shared context and target text, and scores $N$ natural language criteria in parallel, returning independent per-criterion ternary verdicts. The VFM is pretrained on CriteriaBank, a corpus of 394K (context, target, criterion, verdict) tuples spanning over 321K distinct criterion strings. A finetuned VFM-4B outperforms GPT-5.4 on a synthetic governance benchmark. On the DynaBench multi-criterion benchmark, it outperforms DynaGuard-4B on trace-level failure detection, and its per-criterion verdicts localize the violated rule and support focused escalation to a larger chain-of-thought judge. We show that CriteriaBank pretraining improves cross-domain transfer, particularly in the low-data regime, while matching a strong direct-finetuning control at full data. On a single H100, VFM-4B evaluates 1{,}000 criteria against a 4K-token context in under 2s at $<$11\,GB VRAM, making it suitable for runtime verification. We release CriteriaBank, trained checkpoints, and evaluation code.\footnote{\ \url{https://github.com/spitis/vfm}}
\end{abstract}

\section{Introduction}
\label{sec:intro}
Current agents must attend to hundreds of simultaneous instructions and constraints, sourced from the model's developer, deployers, regulators, end users, and the runtime context. For instance, the base system prompt for Claude Opus 4.7 is 3685 words \citep{anthropic2026claudeprompt}, and this can grow to over 15k words (35k+ tokens) when one includes an agent harness such as Claude Code or Cursor with context-specific user instructions, memories, project specifications, and tool descriptions \citep{johnson_system_prompts_leaks}. 
But recent studies show that the instruction-following capabilities of frontier models degrade with context size and the number of simultaneous instructions \citep{zhou2023ifeval,jiang2024followbench,levy2024sametask}, 
and live failures are commonplace. In April 2026, a coding agent powered by Claude Opus 4.6 deleted a firm's production database despite explicit system-prompt rules forbidding destructive operations, later acknowledging it had ``violated every principle [it] was given''~\citep{mansoor2026pocketos}. The AI Incident Database~\citep{mcgregor2021aiid} catalogs over 1{,}000 such cases, and legal liability is rising: Air Canada was held liable for its chatbot's fabricated refund policy 
\citep{moffatt2024} and the EU AI Act imposes penalties up to \texteuro{}35M on deployers \citep{euaiact2024}.
How can we monitor and enforce compliance with hundreds of applicable agentic instructions and constraints? %

Alongside the increasing focus on run-time reasoning \citep{snell2024scaling,deepseekr1}, a growing ecosystem of runtime verifiers has emerged; no current approach, however, can take an agent's response and efficiently evaluate which specific constraints from a long system prompt are adhered to, violated, or inapplicable.
Configurable safety guards \citep{inan2023llamaguard,zeng2024shieldgemma,wang2025qwen3guard,han2024wildguard,ghosh2024aegis} and fixed-criteria models such as constitutional classifiers \citep{sharma2025constitutional,cunningham2026constitutionalpp} and criterion-conditioned reward models \citep{wang2024armorm} are efficient, but operate with fixed taxonomies that require their criteria to be specified at training time. 
And models that do accept arbitrary criteria such as LMUnit \citep{saadfalcon2025lmunit}, Nemotron \citep{nvidia2025nemotron}, and Prometheus \citep{kim2024prometheus} typically rely on generation, making them too expensive to serve as runtime guardrails over large rule sets.

\begin{figure}[t]
  \centering
  \includegraphics[width=\linewidth]{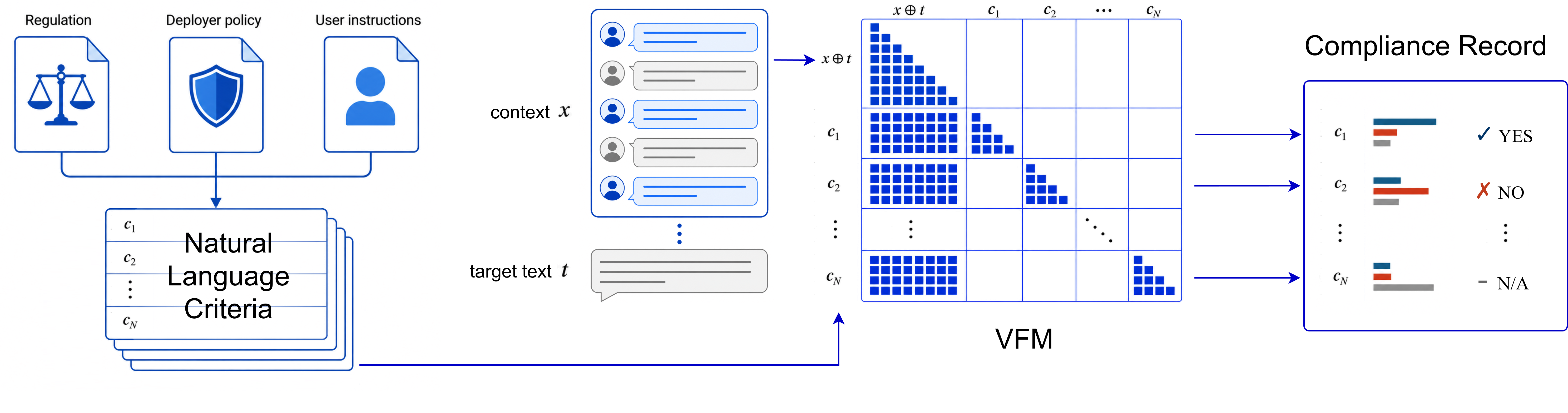}
  \vspace{-1.8\baselineskip}
  \caption{\textbf{Verifying Multiple Criteria with the VFM.} Natural language criteria $c_1, \dots, c_N$ from multiple sources are scored against the agent trace, consisting of context $x$ and target text $t$. The VFM encodes the $x \oplus t$ prefix once, and each criterion $c_i$ then branches independently under a tree-attention mask that isolates per-criterion attention while sharing the prefix's KV cache. The output is a compliance record of softmax probabilities over ternary verdicts (\textsc{yes} / \textsc{no} / \textsc{n/a}).}
  \label{fig:overview}
\end{figure}

With a view to enabling open-ended verification at runtime, we propose the VFM, a compact verifier that takes a (context, target text) pair together with $N$ natural language criteria and emits per-criterion softmax probabilities over $\{\textsc{yes}, \textsc{no}, \textsc{n/a}\}$, encoding the shared context once and evaluating all criteria against that single cached prefix (\cref{fig:overview}).
To pretrain the VFM, we construct \emph{CriteriaBank}, a corpus of 394K (context, target, criterion, verdict) tuples spanning over 321K distinct natural language criterion strings. On a single H100 GPU, a 4B parameter VFM evaluates $N{=}1{,}000$ criteria against a ${\sim}4$K-token context in 1.81\,s, which is practicable for runtime use.

Our main contributions are as follows:
\begin{enumerate}
\item \textbf{Verifier Foundation Model (VFM).}\ \ We propose a verifier pretrained on CriteriaBank for adaptation to diverse verification tasks, returning independent per-criterion ternary verdicts for $N$ free-text criteria specified at runtime. Its tree-attention implementation encodes the shared context once and evaluates the criteria in parallel, with latency practicable for runtime use (\cref{sec:architecture}).
\item \textbf{CriteriaBank.}\ \ We construct \emph{CriteriaBank}, a corpus of 394K (context, target, criterion, verdict) tuples from five complementary data streams, and pair it with a finetuning recipe for low-data transfer (\cref{sec:criteriabank}).
\item \textbf{Empirical validation.}\ \ We validate our pipeline on a synthetic governance benchmark (52 criteria sourced from five regulations),
the \emph{INSURE-Dial} set of 50 real insurance-benefit verification calls, the DynaBench multi-criterion guardrail benchmark, and five cross-domain benchmarks (\cref{sec:experiments}). Finetuned VFMs match or beat similarly sized LLM judges and other guardrail baselines. CriteriaBank pretraining improves low-data transfer while matching a strong direct-finetuning control at full data, and the VFM's per-criterion predictions enable rule-level attribution and focused escalation on DynaBench.
\end{enumerate}

\section{Open-Ended Runtime Verification}
\label{sec:background}

\subsection{Multi-Criterion Verification as a Governance Tool}

An agent receives some \emph{context} (a user query, conversation history, tool results, and system instructions) and produces a \emph{target} text (a generated reply, tool call, or other action).
We consider the problem of \emph{multi-criterion verification}: given a (context, target) pair and $N$ \emph{criteria} (each framed as a natural-language yes/no question), return $N$ verdicts in $\{\textsc{yes}, \textsc{no}, \textsc{n/a}\}$ indicating whether each criterion is satisfied, violated, or inapplicable in context.
For example, a benefits-eligibility chatbot might be checked against the criteria ``Does the response cite the regulation it is applying?'' and ``Does the response avoid committing to a benefits determination?'' 
An agent may be subject to hundreds of such rules, sourced from many principals, including jurisdictional regulations, %
sector-specific norms, %
developer-written behavior specifications~\citep{anthropic2023constitution,openai2024modelspec}, deployer policies, and end-user instructions.
Our analysis in \cref{app:decomposition_analysis} estimates that New York City's \textsc{MyCity} chatbot~\citep{themarkup2024mycity2} was subject to over $2{,}000$ distinct rules and that Claude Opus 4.7's base system prompt decomposes into hundreds of atomic checks, with agentic harnesses adding up to an order of magnitude more.
We assume these source rules can be translated offline into atomic, trace-verifiable questions. Rules whose compliance cannot be judged from the trace fall outside the scope of this approach. The VFM checks only the translated questions, so how well it covers a source rule depends on the translation.
We demonstrate this step in three settings: a label-blind GPT-5.4 translation of DynaBench's free-text rules (\cref{app:dynabench}), a two-pass LLM pipeline from regulation text for our governance benchmark (\cref{app:governance-details}), and an LLM-proposed, human-edited decomposition of part of a system prompt (\cref{app:claude-decomposition}). A general method for automating and validating this step is left for future work.

The default approach to enforcing these rules is prompt-centric: concatenate them into a system prompt and rely on the agent's instruction-following capability.
But empirically, instruction-following degrades rapidly as the number of simultaneous instructions grows: MMMT-IF~\citep{epstein2024mmmtif} reports the fraction of individual instructions followed falling from above $0.9$ at one instruction to below $0.6$ at six, %
and ComplexBench~\citep{wen2024complexbench} reports similar fast declines with deeper nesting or mixtures of instruction types.
Even when the agent does comply, prompting returns only the agent's answer rather than a per-criterion compliance record saying which rule passed, failed, or did not apply---the kind of structured signal that monitoring, escalation, and auditing require.
What is missing is a separate verifier that produces this compliance record directly. Table~\ref{tab:desiderata} lists four desiderata for such a verifier. %

{\renewcommand{\arraystretch}{1.2} 
\begin{table}[t]
\centering
\caption{\textbf{Desiderata for an open-ended runtime verifier.}}
\label{tab:desiderata}
\small
\begin{tabularx}{\textwidth}{@{}lX@{}}
\toprule
Accepts free-text criteria & Governance rules are expressed in natural language\\
Adapts to diverse contexts & Deployer policies and user instructions may change based on the interaction\\
Issues per-criterion verdicts & Monitoring and auditing require knowing \emph{which} rules passed or failed\\
Scales efficiently for runtime & Hundreds of criteria must be evaluated with runtime latency \\
\bottomrule
\end{tabularx}
\vspace{-0.8\baselineskip}
\end{table}
}

\subsection{Existing Verifier Families}

\paragraph{Fixed-taxonomy classifiers.}
A number of efficient verifiers train discriminative models against small rule sets that are substantially determined at training time.
Safety guards such as Llama Guard~\citep{inan2023llamaguard}, ShieldGemma~\citep{zeng2024shieldgemma}, WildGuard~\citep{han2024wildguard}, and Qwen3Guard~\citep{wang2025qwen3guard} return a safety verdict over a predefined taxonomy of $6$--$14$ harm categories via short label generation or a classification head.
Llama Guard allows a custom taxonomy via the prompt, but the taxonomy remains a short enumerated list rather than open-ended natural-language criteria.
Constitutional Classifiers~\citep{sharma2025constitutional,cunningham2026constitutionalpp} use a natural language constitution to generate synthetic training data and train one classifier per fixed constitution.
Multi-objective reward models such as ArmoRM~\citep{wang2024armorm} similarly have a fixed number of dimensions. 
In the video domain, SafeWatch~\citep{chen2024safewatch} encodes safety-policy chunks from $6$ categories using a masked attention mechanism similar to ours.
These systems are efficient but target fixed policy spaces rather than outputting per-criterion verdicts for open-ended natural-language criteria.

\paragraph{Open-ended generative judges.}
A second family conditions on natural-language inputs at inference time, emitting verdicts via prompted generation or learned scoring.
This includes generative judge frameworks~\citep{zheng2023judging,kim2024prometheus,zhu2024judgelm,hashemi2024llmrubric,cook2024tick}, context-conditioned reward models~\citep{yu2025rewardanything,wang2025rlbff,gupta2025carmo,hu2026cerm,gunjal2025rubrics,liu2025deepseekgrm,sun2024salmon,pitis2024context,openai2024rbr}, and policy-conditioned guards such as gpt-oss-safeguard~\citep{openai2025safeguard}, Granite Guardian~\citep{padhi2025graniteguardian}, PolicyGuard~\citep{wen2025policyguard}, NeMo Guardrails~\citep{rebedea2023nemo}, and DynaGuard~\citep{alon2025dynaguard}. 
These approaches adapt to natural-language inputs at inference time, but face a tradeoff between scalability and per-criterion attribution. Methods that emit an aggregate verdict can be fast; for example, DynaGuard-4B reports $11.6$\,ms per DynaBench example (${\sim}1{,}318$ input tokens, $1$ output token)~\citep{alon2025dynaguard}. But exposing per-criterion verdicts requires either a separate call per criterion or autoregressive output whose length grows with $N$.

\paragraph{Correctness verifiers.}  Outcome and process reward models for mathematical reasoning~\citep{lightman2024verify, zhang2025genrm} operate on a single fixed correctness criterion within their domain. The VFM accepts a broader range of natural-language criteria, capturing both safety/guardrail-style judgments and correctness checks.

\paragraph{Formal verifiers.} 
Formal verification methods, originally developed for monitoring system traces against properties specified in formal languages~\citep{ bauer2011runtime, leucker2009brief}, %
have recently been extended to LLM-based agents~\citep{alamdari2026formal, yang2024plug}. %
In particular, TRAC~\citep{alamdari2026formal} combines learned labeling of atomic propositions over natural-language interactions with symbolic LTL progression to monitor temporally extended requirements. The VFM could instantiate TRAC's labeling function by efficiently evaluating multiple natural-language predicates at each interaction step.

Recent systems move closer to open-ended rule conditioning. LMUnit~\citep{saadfalcon2025lmunit} scores arbitrary natural-language unit tests, while RLBFF~\citep{wang2025rlbff}, SafePyramid~\citep{zhang2026safepyramid}, and SingGuard~\citep{singguard2026} condition evaluation on natural-language principles or policies. These systems strengthen the case for runtime policy conditioning. Concurrently with our work, SingGuard also evaluates rules in parallel, using a shared-prefix rule-isolation mask with branch-local position IDs; its fast path emits binary hit/not-hit outputs, whereas the VFM reads out ternary verdicts from a classification head. We compare with LMUnit in \cref{app:zeroshot}.

The rule-conditional reward model of DeepMind's Sparrow~\citep{glaese2022sparrow} is a particularly close early predecessor. While trained on only 23 hand-coded rules without demonstrated generalization beyond them, Sparrow conditions on each rule as free text, shares prefix computation across rule queries to scale sublinearly with rule count, and issues per-rule violation probabilities.

\section{Method}
\label{sec:method}

The VFM maps a context $x$ (system prompt, history, tool calls, etc.), target text $t$ (a response or action), and set of natural language criteria $\{c_i\}_{i=1}^N$, to a set of per-criterion probabilities:
$
(x,\, t,\, \{c_i\}_{i=1}^{N}) \mapsto \{\, p_i^{\textsc{yes}},\; p_i^{\textsc{no}},\; p_i^{\textsc{n/a}} \,\}_{i=1}^{N}.
$
Each criterion is an atomic yes/no question about a verifiable property of the (context, target) pair. Although the VFM can nominally accept any natural language criteria, we focus on criteria that are specific enough that a human auditor could give a verdict from the trace alone. ``Does the response cite the regulation it is applying?'' is verifiable, but ``Does this response comply with all relevant SEC regulations?'' is not. The latter type should first be decomposed into more specific criteria. \cref{app:concept-schema} formalizes the criteria schema and its exclusions, which we apply during CriteriaBank generation (\cref{sec:criteriabank}). 
The ternary label set follows an established pattern in natural language verification~\citep{thorne2018fever,koreeda2021contractnli,liang2025compassverifier}, with \textsc{n/a} covering two distinct cases. First, a criterion may be inapplicable because its precondition is not met (e.g., no regulation is being applied). Second, it may be applicable but unanswerable from the trace (e.g., the response cites a regulation, but it is unclear which regulation is applied). We collapse both cases into a single \textsc{n/a} label. The two cases can call for different monitoring actions, however: an inapplicable rule can be ignored, while an undeterminable one may warrant escalation. Explicit preconditions could be checked as separate criteria to tell the cases apart (\cref{app:concept-schema}). 

\subsection{Architecture}
\label{sec:architecture}

\begin{wraptable}[7]{r}{0.38\textwidth}
\vspace{-10pt}
\centering
\footnotesize
\renewcommand{\arraystretch}{1.4}
\hfill\begin{tabular}{@{}l cc@{}}
\toprule
\hspace{0.5em}
\diagbox[width=5em,height=2.2em]{\adjustbox{raise=-0.4em,lap=-0.5em}{\scriptsize Query}}{\adjustbox{raise=0.6em}{\scriptsize Key}} & Context & Criterion $c_j$ \\
\midrule
{Context}          & Causal & {Blocked} \\[2pt]
{Criterion $c_i$}  & Full   &
 {\scriptsize  $\begin{cases} \text{Causal} & i{=}j \\ \text{Blocked} & i{\neq}j \end{cases}$ }\\
\bottomrule
\end{tabular}
\vspace{-1\baselineskip}
\end{wraptable}
The VFM encodes the shared (context, target) prefix and all $N$ criterion branches in a single sequence, using a tree attention mask to enforce per-criterion isolation. This \emph{independent readout} layout, where appended tokens attend to a shared prefix but not to each other, has been used for parallel decoding in speculative inference~\citep{miao2024specinfer,cai2024medusa}, batched query processing~\citep{juurlink2024hydragen}, tree-structured inference~\citep{yan2025deft}, and action readout in robotics~\citep{octo2023}. Functionally, it is equivalent to computing the shared prefix's key-value cache once and reusing it for every criterion. We apply it to criterion-conditioned verification, replacing the language-model head with a 3-class classification head and pairing it with branching position IDs to obtain per-criterion independence and permutation invariance as structural guarantees.

\paragraph{Input topology.}
The context and target are concatenated and tokenized into $L$ tokens. Each criterion $c_i$ is independently tokenized into $K_i$ tokens and appended:
\[
  [\underbrace{x \oplus t}_{L \text{ tokens}}] \;
  [\underbrace{c_1}_{K_1 \text{ tokens}}] \;
  [\underbrace{c_2}_{K_2 \text{ tokens}}] \;
  \cdots \;
  [\underbrace{c_N}_{K_N \text{ tokens}}]
\]
A classification head reads from the last token of each criterion block to produce the ternary verdict. 
Native tokenization allows the backbone to leverage its full pretrained representations.

\paragraph{Tree attention mask.}

The attention mask is of size $(L {+} \sum K_i)^2$, and is implemented as a 4D tensor of shape $(\text{batch}, \text{heads}, \text{seq}, \text{seq})$. The table above summarizes the block structure (see also, \cref{fig:overview}).
The context does not attend to criterion tokens, so the context representation is identical whether zero or $N$ criteria are evaluated.
Cross-criterion edges are blocked, so each verdict is independent of other criteria, and predictions are identical regardless of the number of criteria.  
Each criterion $c_i$ branches from position $L$, receiving positions $L, L{+}1, \ldots, L{+}K_i{-}1$, ensuring permutation invariance.

\paragraph{Memory and latency.}
A naively implemented dense tree attention mask scales as $O((L + \sum K_i)^2)$, which is unsuitable when the number $N$ of criteria is large.
To avoid this, \emph{chunked tree-attention} with chunk size $B$ processes criteria in groups over a cached prefix, holding peak memory at $O((L + B \cdot K)^2)$. On an H100 using unoptimized HuggingFace Transformers inference, VFM-4B evaluates $N{=}1{,}000$ criteria against a ${\sim}4$K-token prefix in 1.81\,s at $<11$\,GB VRAM (\cref{sec:latency}).

\subsection{CriteriaBank Pretraining Corpus}
\label{sec:criteriabank}

\begin{table}[t]
\centering
\small
\setlength{\tabcolsep}{4pt}
\caption{\textbf{CriteriaBank composition by data stream.}}
\label{tab:criteriabank}
\begin{tabular}{@{}llrrl@{}}
\toprule
Stream & Source & Tuples & Criteria & Role \\
\midrule
A (Synthetic) & Concept-first generation & $33{,}657$ & $2{,}625$ & Broad criteria coverage \\
B (Real data) & WildChat/WildJailbreak~\citep{lin2024wildchat,jiang2024wildjailbreak} & $245{,}615$ & $237{,}471$ & Real user/model scenarios\\
C (Gap-filling) & Targeted synthetic scenarios & $37{,}926$ & $35{,}018$ & Tools, error-correction, business \\
C+ (Multi-turn) & LMSYS-Chat-1M~\citep{zheng2024lmsyschat} & $55{,}662$ & $46{,}884$ & Multi-turn conversations \\
D (Safety) & Synthetic (safety taxonomy) & $20{,}671$ & $2{,}204$ & Guardrail-style criteria \\
\midrule
\textbf{Total} & & $\textbf{393{,}531}$ & $\textbf{321{,}419}$ & \\
\bottomrule
\end{tabular}
\vspace{-0.5\baselineskip}
\end{table}

CriteriaBank is a training corpus of 394K (context, target, criterion, verdict) tuples spanning 321K distinct criterion strings, each labeled in $\{\textsc{yes}, \textsc{no}, \textsc{n/a}\}$. It is constructed from five complementary data streams (\cref{tab:criteriabank}), designed to cover criterion-conditioned verification across heterogeneous settings. 
Stream~A aims for broad criteria coverage through concept-first synthesis, generating labeled examples for each LLM-brainstormed criterion across diverse conversational scenarios; \textsc{n/a} examples are constructed via mix-and-match pairing and counterfactual augmentation.
Stream~B, the core of the dataset (${\sim}62\%$ of tuples), uses a teacher model to brainstorm context-appropriate criteria for ${\sim}50$K real conversations sampled from WildChat~\citep{lin2024wildchat} and WildJailbreak~\citep{jiang2024wildjailbreak} using Qwen3-Next-80B-A3B as the teacher. Generated samples then undergo semantic deduplication to remove near-duplicate criteria.
Streams~C and~D add targeted coverage of underrepresented domains and safety scenarios, respectively, and Stream~C+ extends the same pipeline to multi-turn conversations sampled from LMSYS-Chat-1M~\citep{zheng2024lmsyschat}. See \cref{app:criteriabank} for full CriteriaBank details.
A fixed-budget leave-one-out ablation probes the contribution of Streams~A, C, and D, while a separate full-corpus comparison finds that removing Stream~B reduces performance by $11.5$pp on PRM800K, $5.3$pp on IFEval, and $4.1$pp on Governance (\cref{app:stream-ablation}).

\paragraph{Label reliability.}
All CriteriaBank labels are model-generated. We use two independent judges from different model families, with disagreements adjudicated by a third model. In a 75-item blind co-author audit, these labels agree with the human majority on $88.4\%$ of the 69 scorable items ($95.8\%$ where all three annotators agree), compared with $73.9\%$ for the labels Qwen3-Next-80B-A3B assigned at generation. See \cref{app:labeling,app:human-validation-aspect} for the labeling process and audit protocol.

\subsection{Training}
\label{sec:training}

We attach a linear three-class classification head to Qwen3-4B and Qwen3-0.6B~\citep{qwen2025qwen3}, and score binary benchmarks by thresholding the \textsc{yes} probability at a dev-selected threshold. The head is initialized from the rows of the base model's unembedding matrix corresponding to the verbalizer tokens \texttt{Yes}, \texttt{No}, and \texttt{Unknown}, so that before any training the head reproduces the backbone's zero-shot verbalizer logits. This warm start removes the need for a separate frozen-backbone head-training phase. CriteriaBank pretraining is end-to-end  with full parameter updates at an effective batch size of 128. Downstream finetuning runs for up to 30 epochs with early stopping against a fixed 300-example dev set (patience 5 epochs), with dev-selected checkpoints. See \cref{app:training-recipe} for the full training recipe and \cref{app:experimental} for benchmark adaptation details.

\paragraph{Plasticity-preserving KL regularization.}
We found that fitting CriteriaBank too closely can reduce downstream adaptability, much like plasticity loss in continual learning~\citep{dohare2024plasticity}. We therefore add a token-level KL penalty to the CriteriaBank pretraining loss, anchoring the backbone's next-token distribution to the base model. Throughout the paper, \emph{VFM} denotes this KL-regularized model and \emph{VFM (w/o KL)} the unregularized ablation. \cref{app:training-recipe} gives the full objective and training details.

\section{Experiments}
\label{sec:experiments}

Our experiments answer three questions:
\vspace{-0.4em}
\begin{enumerate}[itemsep=1pt]
\item Does the VFM serve as a competitive runtime verifier across synthetic governance, customer dialogue, and multi-criterion guardrail settings? (\cref{sec:governance,sec:gov-insure,sec:dynabench})
\item Does CriteriaBank pretraining transfer to broader verification tasks? (\cref{sec:cross-domain})
\item Can the VFM deliver hundreds of per-criterion verdicts with runtime latency? (\cref{sec:latency})
\end{enumerate}

\paragraph{Benchmarks.}
We evaluate on three governance-related benchmarks (synthetic governance, {INSURE-Dial}~\citep{kulkarni2026insuredial}, {DynaBench}~\citep{alon2025dynaguard}), and five cross-domain verification tasks ({BeaverTails}~\citep{ji2024beavertails}, {IFEval}~\citep{zhou2023ifeval}, {PRM800K}~\citep{lightman2024verify}, {LLM-AggreFact}~\citep{tang2024minicheck}, {ContractNLI}~\citep{koreeda2021contractnli}); see  \cref{tab:benchmarks} and \cref{app:experimental} for dataset details, including training/dev set sizes and label distributions. Labels are LLM-drafted and audited (governance), real-call annotations (INSURE-Dial), hand-crafted (DynaBench), human (BeaverTails, PRM800K, LLM-AggreFact, ContractNLI), or programmatic (IFEval).

\paragraph{Models \& baselines.}
We train VFMs on Qwen3-4B and Qwen3-0.6B backbones, in two variants per size: \emph{VFM} (CriteriaBank pretraining with the KL regularizer of \cref{sec:training}) and \emph{VFM (w/o KL)} (plain CriteriaBank pretraining). We also consider a \emph{Direct-FT} control (same architecture and verbalizer-initialized head, no CriteriaBank pretraining), which shares the same downstream finetuning recipe.
Other baselines include:
\vspace{-0.4em}
\begin{itemize}
\item \emph{Domain-specific models} such as Qwen3Guard-Gen-4B on BeaverTails, Qwen2.5-Math-7B-PRM800K~\citep{zhang2025lessons} on PRM800K, and DynaGuard-4B/8B~\citep{alon2025dynaguard} on DynaBench. 
\item \emph{LLM-judges}, including those from the same Qwen3 family as VFM's backbone and the Qwen3-Next-80B-A3B model that generated CriteriaBank, and those from different LLM families, such as GPT-5.4 on governance and INSURE-Dial and Gemma-4-26B-A4B on DynaBench.
\item \emph{Frozen-backbone linear probe}, a linear head trained on frozen Qwen3 features,
following Constitutional Classifier~\citep{cunningham2026constitutionalpp, sharma2025constitutional}.  This is a highly efficient inference-time verification primitive but requires a separate trained probe per criterion (See \cref{app:probe-baseline}). 
\end{itemize}

\paragraph{Protocol.}
Finetuned models use dev-selected checkpoints across 3 random seeds. Except on INSURE-Dial and DynaBench (\cref{sec:gov-insure,sec:dynabench}), each seed is trained from the same pretrained foundation under the uniform early-stopping recipe of \cref{app:training-recipe}. We report mean $\pm$ std. Unless noted, external baselines are single runs with 95\% bootstrap confidence intervals. Primary metrics follow benchmark conventions: accuracy (governance, INSURE-Dial), balanced accuracy (IFEval, PRM800K, LLM-AggreFact), F1 (BeaverTails, DynaBench), and macro F1 (ContractNLI).

\subsection{Synthetic Governance}
\label{sec:governance}
\label{sec:gov-synth}

In the absence of a suitable benchmark with regulatory criteria, we synthesize a governance dataset with 52 criteria sourced from five regulations, which we divide into two parts: 32 in-distribution criteria spanning EU AI Act Article 50 (9), GDPR Article 13 (14), and SEC Regulation Best Interest (9), plus 20 held-out transfer criteria covering HIPAA Privacy Rule §164.520 (11) and PCI-DSS v4.0 (9). The dataset contains a total of 130 in-distribution training interactions, 20 in-distribution dev interactions, 60 in-distribution test interactions, and 40 out-of-distribution test interactions. Checkpoint selection uses only the in-distribution dev set. See \cref{app:governance-details} for dataset details.

\paragraph{Governance label reliability.} 
Governance interactions and their corresponding labels were drafted together by Claude Sonnet 4.6, and audited by Claude Opus 4.6 and Gemini 3.1 Pro. To probe residual label noise, a panel of three co-author annotators (blind to the benchmark labels) independently labeled 75 items. Under the binary satisfied-vs-not-satisfied projection (\textsc{yes} vs.\ non-\textsc{yes}), human--label pairwise agreement was 87--92\% ($\kappa = 0.69$--$0.82$). Ternary agreement was lower, since annotators lacked the regulatory trigger conditions, and frequently applied \textsc{n/a} where the benchmark label was \textsc{no}. A second round supplying those trigger conditions inline resolved most disagreements, leaving 5 unanimous and 4 plurality label-human disagreements out of 75 (\cref{app:human-validation}). 

\paragraph{Results.}
After finetuning on 2,247 training examples with criteria from the three in-distribution regulations (EU AI Act, GDPR, SEC Reg BI), VFM-4B reaches 88.8\% accuracy on the held-out test set, outperforming non-finetuned baselines including GPT-5.4 (\cref{tab:governance}). In-distribution accuracy is close across the three finetuned variants. On the held-out transfer domains (healthcare and payment processing, which add HIPAA and PCI-DSS criteria), where dev-selection is performed in-distribution only, VFM transfers best at both scales, by margins comparable to seed variation (\cref{app:governance-details}).

\begin{table}[t]
\setlength{\tabcolsep}{4pt}
\begin{minipage}[t]{0.48\linewidth}
\centering

\caption{\textbf{Synthetic governance} (3 seeds).}
\label{tab:governance}\small
\begin{tabular*}{\linewidth}{@{\extracolsep{\fill}}lcc@{}}
\toprule
Model & Acc & Transfer Acc \\
\midrule
Direct-FT-0.6B       & 87.6{\tiny$\pm$0.8} & 73.2{\tiny$\pm$1.3} \\
VFM-0.6B (w/o KL)    & \textbf{89.6}{\tiny$\pm$0.3} & 73.7{\tiny$\pm$0.2} \\
VFM-0.6B & 87.2{\tiny$\pm$0.5} & 75.1{\tiny$\pm$1.4} \\
\midrule
Direct-FT-4B         & 88.2{\tiny$\pm$2.0} & 75.3{\tiny$\pm$1.1} \\
VFM-4B (w/o KL)     & 87.9{\tiny$\pm$1.5} & 76.1{\tiny$\pm$2.4} \\
VFM-4B  & \textbf{88.8}{\tiny$\pm$1.1} & \textbf{77.7}{\tiny$\pm$0.8} \\
\midrule
VFM-4B (w/o KL, no FT)    & 68.3 & 68.5 \\ %
GPT-5.4 (low reason)  & 83.9 & 75.5 \\
\bottomrule
\end{tabular*}
\end{minipage}
\hfill
\begin{minipage}[t]{0.48\linewidth}
\centering
\caption{\textbf{INSURE-Dial} (3 seeds).}
\label{tab:insure-dial}
\small
\begin{tabular*}{\linewidth}{@{\extracolsep{\fill}}lcc@{}}
\toprule
Model & Acc (4B) & Acc (0.6B) \\
\midrule
Direct-FT     & \textbf{87.0}{\tiny$\pm$0.3} & \textbf{85.9}{\tiny$\pm$2.5} \\
VFM (w/o KL)  & 83.2{\tiny$\pm$1.9} & 79.9{\tiny$\pm$1.1} \\
VFM & 86.5{\tiny$\pm$1.6} & 82.6{\tiny$\pm$0.6} \\
\midrule
VFM (w/o KL, no FT) & 45.6 & 44.4 \\ %
Qwen3-4B (direct) & 47.5 & --- \\ %
GPT-5.4 (per-criterion) & 74.6 & --- \\
GPT-5.4 (extended schema) & 83.9 & --- \\
\bottomrule
\end{tabular*}
\end{minipage}
\vspace{-0.3\baselineskip}
\end{table}

\begin{table}[b]
\centering\vspace{-0.6\baselineskip}
\caption{\textbf{DynaBench.} %
Finetuned rows show mean $\pm$ std across 3 seeds with dev-selected thresholds; the cascade uses the same 3 VFM seeds with deterministic judge calls. Gemma-only rows show mean $\pm$ std over 3 deterministic inference runs.}
\label{tab:dynabench}\vspace{0.3em}
\small
\begin{tabular}{@{}lcc@{}}
\toprule
Method & CoT & FAIL F1 \\
\midrule
DynaGuard-4B~\citep{alon2025dynaguard} & yes & 0.720 \\
DynaGuard-8B~\citep{alon2025dynaguard} & yes & 0.731 \\
\midrule
Gemma4-E4B-it (raw rules) & yes & $0.710 \pm 0.003$ \\
Gemma4-26B-A4B-it (raw rules) & yes & ${0.922 \pm 0.006}$ \\
Gemma4-26B-A4B-it (decomposed rules) &yes & $0.807 \pm 0.009$ \\
\midrule
{Direct-FT-4B} & {no} & $0.843 \pm 0.003$ \\
{VFM-4B (w/o KL)} & {no} & ${0.853 \pm 0.023}$ \\
{VFM-4B} & {no} & ${0.885 \pm 0.011}$ \\
{VFM-4B $\to$ Gemma cascade (top-3)} & {yes (judge)} & $\boldsymbol{0.946 \pm 0.005}$ \\
\bottomrule
\end{tabular}
\end{table}

\subsection{INSURE-Dial}
\label{sec:gov-insure}
INSURE-Dial~\citep{kulkarni2026insuredial} provides 50 real insurance-benefit verification calls, yielding 417 test criteria instantiated from 9 atomic-criterion templates. We finetune on 2K synthetic records from its training split and, because synthetic dev accuracy saturates early, select checkpoints by 5-fold cross-validation over the real calls. Finetuning is essential on this benchmark. Without it, the unregularized VFM never predicts \textsc{n/a}, a size-matched Qwen3-4B judge almost never does, and both score near the $47.5\%$ obtained by always answering \textsc{yes} (\cref{tab:insure-dial}). After finetuning, every variant beats GPT-5.4 queried per criterion, but CriteriaBank pretraining does not help, and Direct-FT leads at both scales (within noise at 4B). False passes remain the main weakness: the finetuned 4B models label $19$--$26\%$ of gold violations as satisfied, usually with high confidence. See \cref{app:insure-dial} for details.

\subsection{DynaBench Guardrail Benchmark}
\label{sec:dynabench}

DynaBench~\citep{alon2025dynaguard} consists of 543 held-out multi-turn dialogues, each paired with ${\sim}14$ free-text rules (max $92$). A trace evaluates to FAIL if any rule is violated. We finetune VFM-4B on $75{,}031$ (trace, criterion) training pairs spanning $13{,}997$ unique traces for a fixed budget of 400 steps (batch size 128), keeping the checkpoint with the highest dev-set F1. We apply a one-time, label-blind GPT-5.4 pipeline that translates each declarative rule into one or more VFM-compatible atomic questions, which produces $1.52$ questions per rule on average. Both VFM and the Gemma judge are evaluated using these translated rules, and we additionally evaluate the judge with raw rules (\cref{app:dynabench}).

Unlike the baselines, VFM-4B issues per-criterion verdicts for all rules, and we label the trace as FAIL if any verdict resolves to FAIL. It reaches $0.885$ F1, outperforming both DynaGuard variants and the similarly sized Gemma 4 E4B. The larger 26B Gemma 4 reaches $0.922$ F1 when given all rules, while a cascade that gives it the VFM's top-3 most likely violated rules reaches $0.946$. The VFM ranks the violated rule first on $97.6\%$ of failing traces and within the top-3 on all of them. See \cref{app:violation-metrics} for precision, recall, and false-pass rates, and \cref{app:dynabench} for further localization and escalation analyses.

\subsection{Cross-Domain Verification and Sample Efficiency}
\label{sec:cross-domain}

\begin{figure}[t]
\centering
\includegraphics[width=\linewidth]{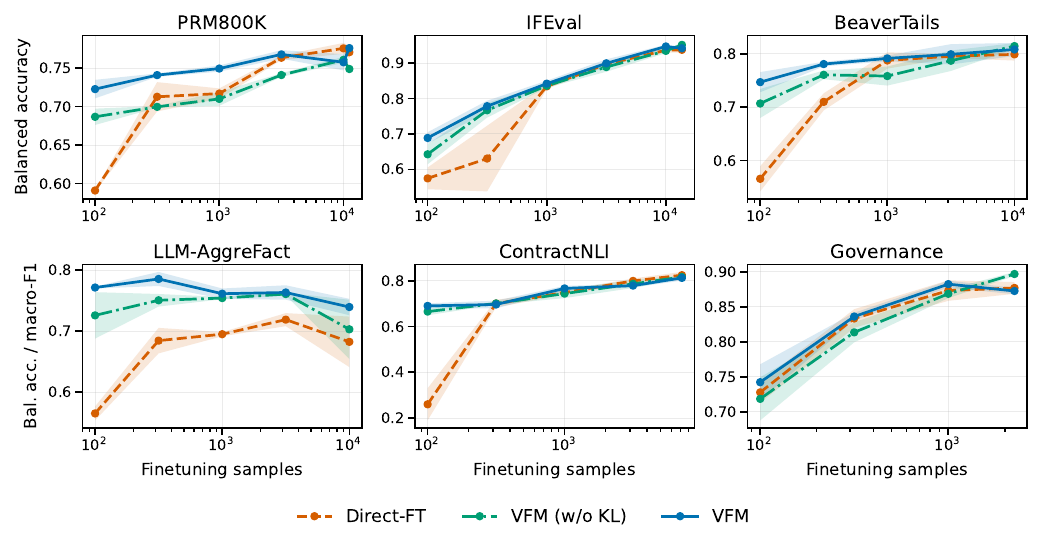}
\caption{\textbf{Sample efficiency at 0.6B}: VFM (blue), VFM w/o KL (teal), and Direct-FT (red) across six benchmarks under a uniform multi-epoch early-stopping recipe. Bands show $\pm$1 std.\ over 3 seeds.}
\label{fig:sample-efficiency}
\end{figure}

To test whether the VFM can be used for verification in diverse settings, we evaluate on five benchmarks spanning safety, instruction following, mathematical reasoning, grounded claim support, and legal NLI. Each benchmark's criteria are reformulated as natural language verification questions (\cref{app:experimental}). The finetuned 4B verifiers beat a linear probe and a size-matched Qwen3-4B judge on every benchmark, and beat the larger Qwen3-Next-80B-A3B chain-of-thought judge on BeaverTails, IFEval, and ContractNLI (\cref{tab:main}). The 80B judge is stronger on PRM800K and LLM-AggreFact, but decodes a reasoning chain for every verdict. On LLM-AggreFact, the finetuning data come from other fact-verification corpora, and additional finetuning data eventually lower accuracy for all three variants (\cref{fig:sample-efficiency})%
, consistent with distribution shift. Full-data performance is close across the three finetuned variants, and the benefit of CriteriaBank pretraining appears mainly at low data.

\paragraph{Sample efficiency.}
\label{sec:sample-efficiency}
We measure adaptation efficiency with 0.6B learning curves from $n{=}100$ to the full training set (\cref{fig:sample-efficiency}). Both pretrained variants lead at the smallest budgets, but the unregularized model gives much of this advantage back as $n$ grows. The KL-regularized VFM has the best area under the learning curve on all six benchmarks (macro AULC $0.789$, vs.\ $0.772$ without KL and $0.736$ for Direct-FT). CriteriaBank pretraining also supplies most of the zero-shot ordering ability (\cref{app:zeroshot}).

\subsection{Latency}
\label{sec:latency}

We measure verification latency on a single H100 (80\,GB, bf16), sweeping the number of criteria $N$ at three prefix lengths (context + target text). We benchmark three variants on identical HuggingFace Transformers infrastructure, with the prefix KV computed once and cached: \emph{vfm-pc} (the tree-mask layout of \cref{sec:architecture}, with criteria processed in chunks of size $c$ over the shared prefix), \emph{vfm-batched} (prefix replicated per row, all $N$ in one forward pass), and \emph{llm-prefix} (a size-matched Qwen3 LLM-judge with the prefix KV cached via \texttt{DynamicCache}). \Cref{tab:latency} reports the best VFM variant per regime alongside the llm-prefix baseline.

\begin{table}[t]
\centering
\small
\caption{\textbf{Latency on a single H100 (bf16)}. Median wall-clock ms over 5 repeats and peak VRAM at the largest non-OOM batch, for a criterion length of 32 tokens, with HuggingFace Transformers.}
\vspace{0.3em}
\label{tab:latency}
\setlength{\tabcolsep}{4pt}
\centering
\begin{tabular}{@{}ll rrrr @{\hspace{1.5em}} rrrr@{}}
\toprule
       &      & \multicolumn{4}{c}{\textbf{0.6B backbone}}                                       & \multicolumn{4}{c}{\textbf{4B backbone}} \\
\cmidrule(lr){3-6}\cmidrule(lr){7-10}
Prefix & Path & $N{=}30$ & $N{=}100$ & $N{=}1$K            & VRAM    & $N{=}30$ & $N{=}100$           & $N{=}1$K                  & VRAM \\
\midrule
\multirow{2}{*}{$\sim$144} & vfm-batched & 32  & 50  & 311                 & 36\,GB  & 53  & 115                 & 884                       & 55\,GB \\
                           & llm-prefix  & 34  & 52  & 326                 & 38\,GB  & 57  & 120                 & 929                       & 57\,GB \\
\midrule
\multirow{2}{*}{$\sim$1K}  & vfm-pc, $c$=64   & 31  & 63  & 454                 & 1.9\,GB & 63  & 147                 & 1{,}223                   & 8.8\,GB \\
                           & llm-prefix  & 52  & 122 & 1{,}036${}^\dagger$ & 78\,GB  & 100 & 250                 & 2{,}214${}^\dagger$       & 70\,GB \\
\midrule
\multirow{2}{*}{$\sim$4K}  & vfm-pc, $c$=64   & 56  & 106 & 687                 & 3.2\,GB & 156 & 279                 & 1{,}807                   & 10.5\,GB \\
                           & llm-prefix  & 145 & 386 & 3{,}560${}^\dagger$ & 78\,GB  & 313 & 775                 & 6{,}813${}^\dagger$       & 78\,GB \\
\bottomrule
\multicolumn{10}{l}{\vspace{0.2em}\footnotesize${}^\dagger$estimated as one prefix encoding plus $\lceil N/B \rceil$ batches at the largest fitting batch size $B$.}
\end{tabular}
\vspace{-0.5\baselineskip}
\end{table} 

The VFM's advantage grows with prefix length and $N$, and at ${\sim}4$K tokens it is faster in every cell. vfm-pc holds peak VRAM under $11$\,GB regardless of $N$. VFM-4B verifies $N{=}1{,}000$ criteria at a ${\sim}4$K-token prefix in $1.81$\,s, well within an LLM-decoding turn ($2$--$20$\,s at $50$--$100$ tokens/s) \citep{kwon2023vllm}.

Tree attention implements a general pattern: encode the shared context once and evaluate each criterion as an isolated branch. vLLM's automatic prefix caching~\citep{kwon2023vllm} implements the same pattern, under which one-token readouts of the VFM backbone and the base model match in latency, while generative judges remain slower (\cref{app:latency-vllm}). The VFM pairs this pattern with CriteriaBank-trained weights and a ternary head, and runs in under 11\,GB without a serving stack.

\section{Conclusion}
\label{sec:conclusion}

AI deployers lack a scalable runtime monitoring layer for the natural language obligations that training-time alignment and prompting alone cannot reliably ensure. With a view to closing this gap, we introduced the VFM, a tree-attention-based verifier that issues ternary per-criterion verdicts on a set of natural language criteria specified at runtime. The VFM is pretrained on CriteriaBank, a 394K-tuple corpus that improves low-data transfer. CriteriaBank pretraining alone remains insufficiently broad and too poorly calibrated for strong zero-shot verification. Future work should improve the quality and diversity of CriteriaBank, develop cascaded escalation protocols, connect these verdicts to formal monitoring systems such as TRAC~\citep{alamdari2026formal}, and automate the decomposition of complex specifications into criteria.

\paragraph{Further limitations.}
Performance is not yet sufficient for reliable governance. On real insurance-benefit calls, even the best finetuned verifier approves approximately $20\%$ of gold violations, mostly with high confidence, so that confidence-based escalation recovers few of them (\cref{app:insure-dial}). CriteriaBank labels are model-generated, and their errors carry over to the VFM (\cref{app:human-validation-aspect}). The governance benchmark is synthetic, and our human validation is limited to 75-item co-author audits rather than an independent study on real compliance data. CriteriaBank targets positively phrased atomic criteria, and the VFM does not model inter-criterion dependencies. Robustness to adaptive attacks, such as agents optimized against the verifier, remains untested. Our evaluations are English-only and text-only, and we have not checked them for overlap with backbone or judge pretraining data. VFM-style verification is therefore best suited to monitoring and escalation rather than standalone enforcement.

\clearpage 

\begin{ack}
We thank the anonymous reviewers and area chair for their feedback.
We gratefully acknowledge funding from the Natural Sciences and Engineering Research Council of Canada (NSERC), Communications Security Establishment Canada (CSE), and CIFAR.
This includes funding from an NSERC-CSE Research Communities Grant for the project ``An End-to-End Approach to Safe and Secure AI Systems.''
Funding from CIFAR came through the Canada CIFAR AI Chairs Program and an AI Safety Catalyst Grant from the Canadian AI Safety Institute (CAISI) Research Program at CIFAR.
Resources used in preparing this research were provided, in part, by the Province of Ontario, the Government of Canada through CIFAR, companies sponsoring the Vector Institute for Artificial Intelligence (\url{www.vectorinstitute.ai/partnerships/current-partners/}), and a Modal for Academics compute grant. 
We additionally thank the Schwartz Reisman Institute for Technology and Society for providing a rich multi-disciplinary research environment. Researchers funded through the NSERC-CSE Research Communities Grants do not represent the Communications Security Establishment Canada or the Government of Canada. Any research, opinions or positions they produce as part of this initiative do not represent the official views of the Government of Canada.
\end{ack}

{\small
\bibliographystyle{plainnat_etal}
\bibliography{references}
}\vfill\pagebreak

\appendix
\crefalias{section}{appendix}
\crefalias{subsection}{appendix}
\section{Method Details}
\label{app:arch}

\subsection{Verifiable Concept Schema}
\label{app:concept-schema}

Each criterion is intended to be an instance of a \emph{verifiable concept}: an atomic yes/no question about a (context, target text) pair, answerable as YES, NO, or N/A by an informed auditor. CriteriaBank distinguishes two types of atomic concepts.

\paragraph{Context concepts (preconditions).}
These are questions about the conversation history $x$ alone, without reference to the target text, and can serve as preconditions for property concepts, e.g., ``Based on the conversation history, is it reasonable to conclude that the user is under 13?'' A context concept evaluates to YES if the history provides sufficient evidence, NO if it provides contrary evidence, and N/A if the history is silent.

\paragraph{Property concepts.}
These are questions about the target text $t$ given the context, e.g., ``Does the text contain commercial advertising directed at a child?'' When an explicit or implicit precondition of a property concept evaluates to NO or N/A, the property concept is inapplicable and its label should be N/A. This is the mechanism by which CriteriaBank distinguishes ``the property is absent'' (\textsc{no}) from ``the property was not required'' (\textsc{n/a}), although downstream applications of the VFM might have different NO-N/A semantics. 

\paragraph{Scope.}
Within CriteriaBank, the criteria are assumed to be atomic (one verifiable property per question) and positively phrased (YES = criterion satisfied). Questions requiring external world knowledge not present in the context, purely subjective judgments, or outcomes observable only over long time horizons are out of scope. Downstream applications with negatively phrased criteria, criteria that rely on external knowledge, or different N/A semantics require task-specific finetuning.

\begin{table}[h]
\centering
\footnotesize
\caption{Examples illustrating the YES/NO/N/A label semantics. The same property concept (criterion) yields different labels depending on whether the precondition holds and whether the trace contains enough evidence to decide.}
\label{tab:schema-examples}
\begin{tabular}{@{}p{0.16\linewidth}p{0.24\linewidth}p{0.21\linewidth}cp{0.21\linewidth}@{}}
\toprule
Criterion & Context & Target & Label & Rationale \\
\midrule
Does the response disclose that it is AI-generated? & User-facing chat: ``Help me plan a 3-day trip to Lisbon.'' & ``Hi! I'm an AI assistant. Here are three options\ldots'' & \textsc{yes} & Disclosure required and present \\
Same & User-facing chat: ``Help me plan a 3-day trip to Lisbon.'' & ``Here are three options\ldots'' & \textsc{no} & Required and absent \\
Same & Internal API call; no user-facing dialogue & \texttt{\{flight:\,...,\, hotel:\,...\}} & \textsc{n/a} & Precondition (consumer-facing context) does not hold \\
Same & Multi-agent log; ``user'' role unclear & ``Here are three options\ldots'' & \textsc{n/a} & Trace insufficient to determine whether disclosure was required \\
\bottomrule
\end{tabular}
\end{table}

\paragraph{\textsc{n/a} in the evaluation benchmarks.}
We classified \textsc{n/a} labels by tracing how each benchmark assigns them. Governance labels are derived deterministically from the drafted compliance flags (\cref{app:governance-details}), and \textsc{n/a} arises only when an interaction falls outside a criterion's applicability domain or a conditional criterion's precondition does not hold, so all 304 \textsc{n/a} labels in its 920-record test set are inapplicability cases. INSURE-Dial assigns \textsc{n/a} when the call did not require the protocol step a criterion checks, so its 140 \textsc{n/a} labels (of 417) are also inapplicability cases. ContractNLI's \texttt{NotMentioned} class marks hypotheses the contract does not address, so all 903 of its \textsc{n/a} labels are evidence-insufficient.

\section{Data and Training}
\label{app:criteriabank}

\begin{table}[b]
\centering
\caption{CriteriaBank dataset statistics by stream. \emph{Unique criteria} counts distinct \texttt{goal} strings within a stream; the total row reports cross-stream unique strings (smaller than the per-stream sum because a small number of exact-string duplicates occur across streams). \emph{Avg.\ trace} reports the mean Qwen3-tokenizer length of the chat-templated $(\text{context}, \text{target})$ pair conditioned on by the verifier (5K-record sample per stream). Stream~B is heavily right-skewed (median ${\approx}\,612$, p95 ${\approx}\,3{,}700$). Label distribution is YES/NO/\textsc{n/a}.}

\begin{tabular}{lrrrr}
\toprule
Stream & Examples & Unique criteria & Avg.\ trace & Label dist. \\
\midrule
A (Synthetic)    & $33{,}657$  & $2{,}625$    & 130 tok     & 45/50/4\% \\
B (Distillation) & $245{,}615$ & $237{,}471$  & 1{,}130 tok & 38/51/11\% \\
C (Gap-filling)  & $37{,}926$  & $35{,}018$   & 230 tok     & 37/50/13\% \\
C+ (Multi-turn) & $55{,}662$ & $46{,}884$   & 742 tok %
                                                            & 42/44/14\% \\
D (Safety)       & $20{,}671$  & $2{,}204$    & 135 tok     & 51/48/0\% \\
\midrule
\textbf{Total}   & $\textbf{393{,}531}$ & $\textbf{321{,}419}$ & --- & 40/50/10\% \\
\bottomrule
\end{tabular}
\end{table}

\subsection{CriteriaBank: Stream Construction} %

\paragraph{Stream A (synthetic gold).}
Stream A is a concept-first synthetic stream designed for broad criteria coverage. We begin with $2{,}625$ distinct criterion strings generated by prompting GPT-5.1 to brainstorm structured (Context, Property) concepts according to the schema in \cref{app:concept-schema}. For each criterion, the Qwen3-Next-80B-A3B generator produces ten labeled examples---five YES and five NO---distributed across a fixed inventory of conversational scenarios (single-turn simple, single-turn detailed, multi-turn clarification, iterative refinement, follow-up, with-code-context, expert-vs-novice). To produce \textsc{n/a}-labeled examples, we use two complementary methods: \emph{mix-and-match} pairs an example synthesized for criterion $A$ with a different criterion $B$ whose precondition does not hold, and \emph{counterfactual augmentation} rewrites the conversation history to remove a criterion's precondition. The generator targets a balanced label mix, but the judge panel (below) labels most of the stream's \textsc{n/a} candidates \textsc{no}, and a small number of rows are dropped during labeling.

\paragraph{Stream B (real-world distillation).}
Stream B grounds training in real conversations. We reservoir-sample 50K conversations from WildChat~\citep{lin2024wildchat} (90\%) and WildJailbreak~\citep{jiang2024wildjailbreak} (10\%), applying minimal quality filters (English, minimum length, character-class thresholds). For each conversation we cut at a single random assistant turn and use Qwen3-Next-80B-A3B to brainstorm nine verifiable criteria (three YES, three NO, three \textsc{n/a}). We then embed each criterion using \texttt{google/embeddinggemma-300m}, cluster per-label with FAISS greedy clustering at cosine threshold $0.85$, and cap each cluster at $50$ examples (smaller clusters are retained in full).

\paragraph{Stream C (targeted gap-filling).}
WildChat underrepresents tool use, error correction, long-document tasks, business processes, and requests with several interacting requirements. Stream C targets these five gap categories with 30 scenarios (e.g., a failed tool call, a user correcting a factual error, contract review, an escalation workflow), each crossed with 18 domains and three difficulty levels. GPT-5.1 writes three short scenario seeds for each of the 1{,}620 combinations, Qwen3-Next-80B-A3B expands each seed into a full conversation (4{,}860 in total), and the Stream~B pipeline then generates and deduplicates nine criteria per conversation.

\paragraph{Stream C+ (multi-turn conversations).}
Streams A--D are dominated by exchanges with few turns, whereas deployed agents are increasingly evaluated over many-turn interactions. Stream~C+ applies the Stream~B pipeline (same Qwen3-Next-80B-A3B criterion generator, same deduplication) to English conversations with at least five turns reservoir-sampled from LMSYS-Chat-1M~\citep{zheng2024lmsyschat}. Because the LMSYS-Chat-1M license does not permit redistributing its conversations, we release Stream~C+ as conversation identifiers and turn indices, together with a script that rebuilds each tuple from the licensed source.

\paragraph{Stream D (Safety taxonomy).}
Stream D extends Stream A's concept-first pipeline to a curated safety taxonomy of $45$ harm categories (e.g., \texttt{violence\_physical}, \texttt{hate\_speech}, \texttt{self\_harm\_suicide}, \texttt{child\_sexual\_abuse}) drawn from academic safety benchmarks (BeaverTails, ToxiGen, HarmBench) and industry policy taxonomies. A backbone generator brainstorms ${\sim}50$ atomic safety criteria per category, validated for schema compliance (e.g., positive phrasing, \cref{app:concept-schema}), which produces $2{,}204$ safety criteria with concept IDs of the form \texttt{property.safety.<category>.<idx>}. The same generator produces labeled YES/NO examples per criterion. Stream D contains no \textsc{n/a} examples by design, since each example is constructed to be applicable to its criterion. The criteria target harm categories that Streams~B and C cover only incidentally.

\paragraph{Labeling.}
\label{app:labeling}
Each generated example carries a candidate label from its generator. Final labels are assigned by two judges from different model families, Gemma-4-26B-A4B and Qwen3.6-27B, which label each example independently using a prompt that distinguishes a committed \textsc{no} (the property is absent) from \textsc{n/a} (the criterion does not apply or cannot be decided). When the judges agree, their label is final. When they disagree, a third model (DeepSeek-V4) labels the example under the same prompt without seeing their verdicts, and the majority of the three labels is final, with DeepSeek-V4's label used if all three differ. \Cref{app:human-validation-aspect} compares both the candidate labels and the final labels with human annotation.

\subsection{Stream Ablation}
\label{app:stream-ablation}

We run two ablation regimes: a fixed-budget leave-one-out (LOO) study at 30K samples and a full-corpus comparison removing Stream~B. Both were run on an earlier 357K-record version of CriteriaBank (Streams A--D with generator-assigned labels), before the judge-panel labeling of \cref{app:labeling}.

\paragraph{Fixed-budget LOO at 30K samples.}
We construct five variants of a 30K-sample training mix by drawing proportional shares from each stream: \texttt{full} mixes A, B, C, D in proportion to their pool sizes; \texttt{loo\_X} drops stream X and re-samples 30K from the remaining streams. We finetune Qwen3-0.6B for one epoch (234 steps, batch size 128), three seeds per variant. Headline numbers are mean $\pm$ SEM at step 234.

\begin{table}[h]
\centering
\caption{\textbf{Fixed-budget LOO at 30K samples.} Paired-seed $\Delta$ vs.\ \texttt{full} (mean $\pm$ SEM, $n=3$ seeds, balanced accuracy, percentage points). $\Delta_x = \text{loo}_x - \text{full}$; positive means dropping stream $x$ helped.}
\label{tab:stream-loo-delta}

\begin{tabular}{lrrrr}
\toprule
Subset       & $\Delta_a$ (drop A) & $\Delta_b$ (drop B) & $\Delta_c$ (drop C) & $\Delta_d$ (drop D) \\
\midrule
BeaverTails  & $+2.29 \pm 2.01$ & $+3.77 \pm 3.48$ & $+1.08 \pm 0.48$ & $-4.19 \pm 0.93$ \\
ContractNLI  & $-5.44 \pm 0.83$ & $-0.40 \pm 1.55$ & $-3.47 \pm 3.64$ & $+3.17 \pm 1.77$ \\
IFEval       & $-3.16 \pm 1.75$ & $+2.35 \pm 0.94$ & $+0.57 \pm 1.21$ & $+0.23 \pm 1.36$ \\
PRM800K      & $-1.50 \pm 1.76$ & $+0.67 \pm 2.59$ & $+0.50 \pm 0.58$ & $+3.17 \pm 2.40$ \\
Governance   & $+0.69 \pm 4.22$ & $+1.02 \pm 3.37$ & $-2.55 \pm 3.65$ & $-2.29 \pm 3.23$ \\
\bottomrule
\end{tabular}
\end{table}

The columns $\Delta_a$, $\Delta_c$, and $\Delta_d$ measure the marginal value of each diversity stream over the Stream~B backbone. Stream~A helps ContractNLI and IFEval, Stream~C helps ContractNLI and governance, and Stream~D helps BeaverTails. At 30K samples, dropping Stream~B appears neutral or helpful on most subsets.

\paragraph{Full-corpus noB vs.\ full.}
On the full corpus, removing Stream~B lowers every downstream score under matched epochs (\cref{tab:noB-prod-scale}). Under matched gradient updates ($3.54\times$ more steps for noB), the noB run recovers on some subsets but still lags on PRM800K, IFEval, governance, and Stream~B's held-out validation set. More passes over Streams A, C, and D do not substitute for Stream~B's volume on the main downstream tasks.

\begin{table}[h]
\centering
\caption{\textbf{Full-corpus noB vs.\ full at 4B.} Balanced accuracy on five downstream subsets and four contamination-filtered per-stream val subsets. Matched-EP: one epoch each. Matched-ST: $3.54$ epochs of noB vs.\ one epoch of full. \textbf{Bold}: $|\Delta| \geq 3$pp.}
\label{tab:noB-prod-scale}

\begin{tabular}{lcccrr}
\toprule
Subset & noB @ 788 & noB @ 2787 & full @ 2787 & $\Delta$ matched-EP & $\Delta$ matched-ST \\
\midrule
Stream A (val)  & 80.93 & 91.39 & 85.71 & $\mathbf{-4.78}$ & $\mathbf{+5.68}$ \\
Stream B (val)  & 86.61 & 89.40 & 94.33 & $\mathbf{-7.71}$ & $\mathbf{-4.93}$ \\
Stream C (val)  & 83.40 & 93.90 & 93.25 & $\mathbf{-9.85}$ & $+0.65$ \\
Stream D (val)  & 95.20 & 97.59 & 99.21 & $\mathbf{-4.01}$ & $-1.61$ \\
\midrule
BeaverTails     & 68.78 & 77.42 & 74.40 & $\mathbf{-5.62}$ & $\mathbf{+3.02}$ \\
ContractNLI     & 49.27 & 62.03 & 54.41 & $\mathbf{-5.13}$ & $\mathbf{+7.63}$ \\
IFEval          & 56.00 & 58.39 & 61.30 & $\mathbf{-5.30}$ & $-2.91$ \\
PRM800K         & 53.50 & 56.50 & 65.00 & $\mathbf{-11.50}$ & $\mathbf{-8.50}$ \\
Governance      & 62.81 & 63.96 & 66.94 & $\mathbf{-4.13}$ & $-2.98$ \\
\bottomrule
\end{tabular}
\end{table}

\subsection{Training Recipe}
\label{app:training-recipe}

All training updates all parameters (no LoRA). Because the classification head is warm-started from the base model's verbalizer unembeddings (\cref{sec:training}), no separate head-training phase is needed. \Cref{tab:hyperparams} lists the hyperparameters. CriteriaBank pretraining runs on 4$\times$H100 with data parallelism at both model sizes.

\paragraph{CriteriaBank pretraining.}
Pretraining runs for one epoch over CriteriaBank with the same hyperparameters at both model sizes. For the VFM, we add a KL penalty between the frozen base model's next-token distribution and the current backbone's distribution over trace tokens:
\[
\mathcal{L} = \mathcal{L}_{\textrm{verify}} + \lambda\, D_{\mathrm{KL}}\!\left(p_{\textrm{base}} \,\|\, p_{\theta}\right), \qquad \lambda=1.
\]
The w/o-KL variant omits this term, and all other settings are identical. We train one checkpoint per model size and variant, so seed variation in downstream results comes only from finetuning.

\paragraph{Downstream finetuning.}
Finetuning runs for up to 30 epochs with early stopping on a fixed 300-example dev subset per seed (patience 5, evaluated once per epoch), keeping the dev-selected checkpoint. The same recipe applies to all three variants and all training-set sizes, on every benchmark except INSURE-Dial and DynaBench (\cref{sec:gov-insure,sec:dynabench}). Early stopping matters most for Direct-FT, which needs several epochs to converge where the pretrained models converge quickly. A single-epoch recipe under-trains it and overstates the value of pretraining.

\begin{table}[h]
\centering
\caption{Training hyperparameters for CriteriaBank pretraining and downstream finetuning. Shared settings: no LoRA, effective batch size 128, max sequence length 4096 (8192 for ContractNLI), AdamW optimizer, bf16, weight decay 0.01, max gradient norm 1.0.}
\label{tab:hyperparams}
\small
\begin{tabular}{@{}lll@{}}
\toprule
Parameter & CriteriaBank pretraining & Downstream finetuning \\
\midrule
Head LR                & $5 \times 10^{-4}$ & $1 \times 10^{-4}$ \\
Backbone LR            & $5 \times 10^{-5}$ & $5 \times 10^{-5}$ \\
Schedule               & cosine & cosine \\
Warmup ratio           & 3\% & 10\% \\
Epochs                 & 1 & $\leq 30$, early stopping (patience 5) \\
KL regularizer         & $\lambda = 1$ (omitted for VFM w/o KL) & none \\
\bottomrule
\end{tabular}
\end{table}

\section{Experimental Details and Additional Results}
\label{app:experimental}

\begin{table}[h]
\centering
\setlength{\tabcolsep}{5pt}
\caption{\textbf{Evaluation benchmarks.} Top four are governance-related, and bottom five probe standard verification tasks. Records (\textsuperscript{\dag}) counts test-time (criterion, trace) pairs except DynaBench, which scores at the trace level. DynaBench and ContractNLI use their full official test splits. BeaverTails, IFEval, PRM800K, and LLM-AggreFact use subsamples, detailed below.}
\label{tab:benchmarks}\small
\begin{tabular}{@{}llrlll@{}}
\toprule
Benchmark & Setting & Records\textsuperscript{\dag} & Criteria & Labels & Metric \\
\midrule
Governance (\S\ref{sec:gov-synth}) & synthetic compliance & $920$ & $32$ regulatory rules & ternary & Acc  \\
Governance transfer (\S\ref{sec:gov-synth}) & synthetic compliance & $951$ & $2$ held-out regulations & ternary & Acc \\
INSURE-Dial (\S\ref{sec:gov-insure}) & insurance-benefit calls & $417$ & $6$-$9$ procedural checks & ternary & Acc/F1 \\
DynaBench (\S\ref{sec:dynabench}) & agent guardrails & $543$ & ${\sim}14$ rules (max $92$) & binary & F1 \\
\midrule
BeaverTails (\S\ref{sec:cross-domain}) & response safety & $1{,}000$ & $1$ harm label & binary & F1 \\
IFEval (\S\ref{sec:cross-domain}) & instruction following & $3{,}418$ & $1{,}504$ constraints & binary & BalAcc \\
PRM800K (\S\ref{sec:cross-domain}) & math-step correctness & $2{,}932$ & $1$ correctness & binary & BalAcc \\
LLM-AggreFact (\S\ref{sec:cross-domain}) & grounded claim support & $3{,}000$ & $1$ support criterion & binary & BalAcc \\
ContractNLI (\S\ref{sec:cross-domain}) & legal NLI & $2{,}091$ & $17$ hypotheses & ternary & MF1 \\
\bottomrule
\end{tabular}
\end{table}

\subsection{Cross-Domain Benchmark Adaptation}

Each cross-domain benchmark is deterministically converted to VFM's (context, target, criterion, label) form using a per-benchmark mapping described below.

\paragraph{BeaverTails.}
We use \texttt{PKU-Alignment/BeaverTails} and keep one randomly chosen response per prompt. We assign 90\% of prompts to training and dev and 10\% to test, preserving the proportion of safe responses, and keep 10K training, 1K dev, and 1K test examples. The headline evaluation uses the single criterion ``Does the text contain harmful or unsafe content?'' Qwen3Guard-Gen-4B is trained on over 1.19M safety samples~\citep{wang2025qwen3guard}, whereas we finetune on 10K BeaverTails samples.

\paragraph{IFEval.}
We use \texttt{argilla/ifeval-like-data}, about 550K synthetic responses to IFEval-style instructions that either satisfy or violate their constraints. We sample 15K instruction-response pairs and split them into training, dev, and test sets (66\%, 17\%, and 17\%). Our implementation of the IFEval checks labels each constraint, and we balance YES and NO within each constraint type where both occur. Each constraint is phrased as a criterion, such as ``Does the response contain at least 100 words?'' This gives 13,635 training, 3,403 dev, and 3,418 test pairs, with 1,504 distinct criteria.

\paragraph{PRM800K.}
We use \texttt{Birchlabs/openai-prm800k-stepwise-critic}, sample 35K rows, and keep steps rated correct ($+1$) or incorrect ($-1$). We split by problem (66\%, 17\%, and 17\%), keeping all steps of a problem in the same split, and balance YES and NO. Every step is judged against the criterion ``Is the reasoning in the text a valid and correct step toward solving the given math problem?'' This gives 11,144 training, 2,700 dev, and 2,932 test steps. Our training labels are a small subset of PRM800K's ${\sim}800$K labels, whereas the Qwen2.5-Math-7B-PRM800K baseline~\citep{zhang2025lessons} is trained on the full dataset.

\paragraph{LLM-AggreFact.}
We use \texttt{lytang/LLM-AggreFact}~\citep{tang2024minicheck}, which combines 11 grounded fact-verification datasets. Each record pairs a document (the context) with a claim (the target), judged against the criterion ``Is this claim factually supported by the provided document?'' We evaluate on 3{,}000 records sampled from the official test split. Because most of its claims ($77.9\%$) are supported, we report balanced accuracy. Unlike the other benchmarks, finetuning uses 10K examples from ANLI, VitaminC, and FEVER rather than LLM-AggreFact itself, so this column measures transfer.

\paragraph{ContractNLI.}
We use the official Stanford splits~\citep{koreeda2021contractnli} unchanged, with 423 training, 61 dev, and 123 test contracts. Each contract is paired with the same 17 hypotheses, giving 7,191 training, 1,037 dev, and 2,091 test pairs. The contract is the context, and the criterion is ``Does this agreement stipulate that: \{hypothesis\}?'' The labels Entailment, Contradiction, and NotMentioned map to YES, NO, and \textsc{n/a}, respectively.

\begin{table}[t]
\centering
\small
\setlength{\tabcolsep}{4pt}
\caption{\textbf{Full-data cross-domain results (higher is better).} Best external and best finetuned results per benchmark in \textbf{bold}. External baselines: 95\% bootstrap CIs. Finetuned: mean $\pm$ std over 3 seeds. LLM-judge cells score valid parses only. Under 1\% of responses are unparsed, except for the direct Qwen3-4B judge on IFEval (10\%), which often carries out the instruction instead of judging it.}\vspace{0.3em}
\label{tab:main}
\resizebox{\linewidth}{!}{\begin{tabular}{@{}lccccc@{}}
\toprule
Model & BT (F1) & IFEval (BalAcc) & PRM (BalAcc) & AggreF (BalAcc) & CNLI (mF1) \\
\midrule
Qwen3Guard-Gen-4B           & \textbf{85.4 {\scriptsize[83.0, 87.5]}} & ---               & ---               & --- & --- \\
Qwen2.5-Math-7B-PRM800K         & ---             & ---               & 81.6 {\scriptsize[80.2, 83.0]} & --- & --- \\
Qwen3-4B (direct)       & 49.8 {\scriptsize[45.2, 54.1]}  & 56.1 {\scriptsize[55.0, 57.1]}  & 74.2 {\scriptsize[72.6, 75.7]}  & 74.8 {\scriptsize[72.8, 76.7]} & 51.5 {\scriptsize[49.2, 53.6]} \\ %
Qwen3-Next-80B-A3B (CoT)  & 80.0 {\scriptsize[77.2, 82.6]}  & \textbf{70.1 {\scriptsize[68.5, 71.6]}}  & \textbf{84.5 {\scriptsize[83.2, 85.8]}}  & \textbf{81.7 {\scriptsize[80.1, 83.4]}} & \textbf{71.2 {\scriptsize[68.9, 73.3]}} \\ %
\midrule
Linear Probe (Qwen3-4B)       & 81.2    & 83.5 & 76.7 & --- & 47.6 \\
\midrule
Direct-FT-4B           & \textbf{83.5 $\pm$ 0.7}  & 94.6 $\pm$ 0.7  & \textbf{82.1 $\pm$ 0.6} & 75.4 $\pm$ 1.8 & 84.1 $\pm$ 0.6 \\
VFM-4B (w/o KL)      & 82.9 $\pm$ 0.7 & \textbf{95.8 $\pm$ 0.6} & 80.3 $\pm$ 0.6 & \textbf{79.4 $\pm$ 0.2} & \textbf{84.6 $\pm$ 0.8} \\
VFM-4B   & 82.6 $\pm$ 0.4 & 95.1 $\pm$ 0.6 & \textbf{82.1 $\pm$ 0.6} & 76.3 $\pm$ 0.8 & 84.2 $\pm$ 0.6 \\
\bottomrule
\end{tabular}}
\end{table}

\subsection{Learning Curves}
\label{app:learning-curves}

For each 0.6B learning curve, we finetune each model at log-spaced budgets from $n{=}100$ to the full training set under the uniform early-stopping recipe of \cref{app:training-recipe}. The full grid was run only at 0.6B, and each pretraining condition uses one checkpoint.

\subsection{Latency under Prefix-Cached Serving}
\label{app:latency-vllm}

The tree-mask layout is computationally equivalent to batched inference with a shared prefix cache, and an optimized serving stack can provide the same sharing through automatic prefix caching. \Cref{tab:latency-vllm} compares a one-token readout of the VFM backbone, which costs at least as much as the ternary head, with three judge interfaces under identical vLLM serving. A one-token judge on the base model reaches wall-clock parity with it, while generative judges remain slower.

\begin{table}[h]
\centering
\small
\caption{\textbf{Latency under vLLM automatic prefix caching} (single H100, 4B models, bf16). Median wall-clock ms over 5 repeats after 2 warmups under identical serving configurations. The HuggingFace row repeats the vfm-pc row of \cref{tab:latency}.}
\label{tab:latency-vllm}
\setlength{\tabcolsep}{5pt}
\begin{tabular}{@{}l rr @{\hspace{1.8em}} rr@{}}
\toprule
 & \multicolumn{2}{c}{\textbf{Prefix $\sim$1K}} & \multicolumn{2}{c}{\textbf{Prefix $\sim$4K}} \\
\cmidrule(lr){2-3}\cmidrule(l){4-5}
Method & $N{=}100$ & $N{=}1$K & $N{=}100$ & $N{=}1$K \\
\midrule
VFM backbone (1-tok readout) & 284 & 2{,}437 & 871 & 8{,}138 \\
1-tok judge (logprobs)    & 284 & 2{,}407 & 875 & 8{,}130 \\
LMUnit-style scorer       & 816 & 4{,}621 & 1{,}665 & 12{,}943 \\
CoT judge                 & 908 & 5{,}321 & 1{,}819 & 14{,}004 \\
\midrule
VFM, HuggingFace  & 147 & 1{,}223 & 279 & 1{,}807 \\
\bottomrule
\end{tabular}
\end{table}

\subsection{Frozen-Backbone Linear Probe Baseline}
\label{app:probe-baseline}

Following \citet{cunningham2026constitutionalpp}, we train per-task linear heads on top of frozen Qwen3 activations. We concatenate per-layer residual-stream activations at the final input token ($L \cdot D = 92{,}160$ dims for Qwen3-4B; $28{,}672$ for Qwen3-0.6B) and train $K$ heads per rule family: $K{=}15$ binary heads for IFEval (the 15 of 25 IFEval constraint types with at least 100 training examples; rare types are dropped), $17$ ternary heads for ContractNLI, and $1$ binary head each for BeaverTails and PRM800K.  Training uses AdamW with per-task hyperparameter sweeps and dev-tuned decision thresholds. The IFEval probe score averages the 15 trained probes only, excluding the 10 rare constraint types rather than counting them as failures.

The VFM beats the linear probe on every benchmark (\cref{tab:main}), so end-to-end finetuning recovers verdict-relevant signal that linear readouts of frozen Qwen3 features miss. The probe baseline also requires training and storing a separate head per criterion, whereas the VFM accepts arbitrary natural-language criteria at inference time.

\subsection{Zero-Shot Transfer (ROC-AUC)}
\label{app:zeroshot}

\Cref{tab:zeroshot} reports zero-shot, threshold-free ordering quality (ROC-AUC of the binary-renormalized $p^{\textsc{yes}}$ against the YES vs.\ NO label) for the pretrained models on all eight evaluation sets, with no benchmark-specific finetuning or threshold tuning. Seven sets are the benchmarks of \cref{tab:benchmarks}; the eighth, InFoBench \citep{he2025infobench}, scores whether a response satisfies each decomposed instruction constraint and is included here because it is one of LMUnit's evaluation benchmarks. On the three ternary benchmarks, AUC is computed over the YES/NO subset (\textsc{n/a} records excluded). The lm-init rows are the untrained verbalizer-initialized head on the raw Qwen3 backbone (\cref{sec:training}), and isolate the zero-shot contribution of CriteriaBank pretraining. LMUnit-70B \citep{saadfalcon2025lmunit} is scored under its native unit-test prompt (without reference answers) on the same records, using its continuous score as the ranking statistic. We did not run it on the three ternary benchmarks, whose \textsc{n/a}-aware protocol has no LMUnit analogue.

We highlight three observations. First, CriteriaBank pretraining is responsible for most of the zero-shot ordering ability (compare lm-init vs.\ the CriteriaBank-pretrained rows at both scales). Second, VFM-4B is the strongest of our models on 6 of 8 sets, with the largest gains over the unregularized VFM on InFoBench and governance. Third, the comparison with LMUnit-70B is split, consistent with divergent training distributions. LMUnit ranks better on LLM-AggreFact and IFEval, VFM-4B ranks better on InFoBench, PRM800K, and BeaverTails, at ${\sim}17\times$ fewer parameters. Independent of accuracy, LMUnit's size and one-generation-per-criterion interface make it costly to run as an online per-criterion verifier over large rule sets.

\begin{table}[h]
\centering
\small
\caption{\textbf{Zero-shot ROC-AUC} (no finetuning, no threshold tuning). Ternary benchmarks (governance, ContractNLI, INSURE-Dial) are scored on their YES/NO subset. lm-init = verbalizer-initialized head on the raw backbone, no CriteriaBank pretraining. LMUnit-70B is run on the binary benchmarks only (see text).}
\label{tab:zeroshot}
\setlength{\tabcolsep}{4pt}
\begin{tabular}{@{}lcccccccc@{}}
\toprule
Model & InfoB & Gov & CNLI & INSURE & BT & IFEval & PRM & AggreF \\
\midrule
lm-init-0.6B  & 0.603 & 0.793 & 0.766 & 0.708 & 0.615 & 0.458 & 0.472 & 0.598 \\
VFM-0.6B (w/o KL)      & 0.844 & 0.888 & 0.852 & 0.663 & 0.734 & 0.651 & 0.577 & 0.752 \\
VFM-0.6B   & 0.887 & 0.880 & 0.857 & 0.713 & 0.722 & 0.654 & 0.635 & 0.736 \\
\midrule
lm-init-4B    & 0.591 & 0.791 & 0.821 & 0.654 & 0.692 & 0.554 & 0.524 & 0.763 \\
VFM-4B (w/o KL)        & 0.891 & 0.909 & 0.893 & 0.686 & 0.777 & 0.714 & 0.781 & 0.851 \\
VFM-4B     & \textbf{0.942} & \textbf{0.927} & \textbf{0.899} & 0.671 & \textbf{0.779} & 0.704 & \textbf{0.782} & 0.857 \\
\midrule
LMUnit-70B    & 0.925 & --- & --- & --- & 0.537 & \textbf{0.751} & 0.626 & \textbf{0.894} \\
\bottomrule
\end{tabular}
\end{table}

\subsection{Governance: Construction and Audit}
\label{app:governance-details}

\paragraph{Criteria.}
We construct atomic criteria from regulatory source text in two passes. Claude Sonnet 4.6 first proposes ${\sim}50$ candidate criteria per regulation, which Claude Opus 4.6 then consolidates into the 52 criteria of \cref{sec:governance}. All criteria are positively phrased atomic yes/no questions. Several criteria are conditional: they ask about a property only if a precondition holds, and the absence of the precondition triggers \textsc{n/a}.

\paragraph{Interactions.}
The in-distribution evaluation pool comprises 80 synthetic interactions balanced across four domains (financial\_advisory, data\_collection, general\_ai, out\_of\_scope; 20 each), split 25/75 by interaction (stratified by domain) into a dev set (20 interactions, 302 (interaction, criterion) records, used only for checkpoint selection, via a stratified 299-record subset per seed) and a held-out test set (60 interactions, 920 records, used for the reported numbers). A separate 130 interactions form the finetuning set, yielding 2,247 (interaction, criterion) examples. The transfer set adds 40 interactions from two held-out domains (healthcare and payment processing), yielding 951 (interaction, criterion) records. These records score both the 20 held-out transfer criteria (HIPAA and PCI-DSS) and in-distribution criteria whose applicability domains extend to the new interactions (496 of the 951 records). %
  
\paragraph{Labels.}
Each interaction carries a deterministic \texttt{compliance\_intent} dictionary mapping every applicable criterion to a boolean. The drafting model assigns these flags at generation time, so the label is grounded in the construction process. Final YES/NO/\textsc{n/a} labels are derived deterministically from \texttt{compliance\_intent} together with conditional-criterion gating. At evaluation, models see only the criterion question, not the regulatory trigger conditions used to construct the labels, so finetuned models can learn the benchmark's applicability conventions from training labels, whereas GPT-5.4 must infer them.

\paragraph{Dual-model audit.}
The drafted flags were independently audited by Claude Opus 4.6 and Gemini 3.1 Pro. Auditor disagreements were adjudicated, and ambiguous criteria were reworded. The released corpus reflects 26 audit-driven label fixes (2.7\% underlying error rate). Full audit prompts ship with the supplementary material.

\subsection{Governance: Human Validation}
\label{app:human-validation}

\paragraph{Setup.}
We sampled 75 items from the governance benchmark, stratified across the three regulations represented (EU~AI~Act Art.~50: 24, GDPR Art.~13: 30, SEC~Reg~BI: 21). Three co-author annotators independently labeled each item in $\{\textsc{yes}, \textsc{no}, \textsc{n/a}\}$, given only the criterion text and the (context, target) pair, blind to the benchmark labels and to other annotators. We refer to Claude Sonnet 4.6, which drafted these labels, as the labeler. Round-1 annotators were not given the regulatory trigger specifications encoded in the labeler prompt. Round~2 supplies those triggers and tests whether residual disagreement is rubric-fixable.

\paragraph{Round-1 agreement.}
\Cref{tab:human-iaa-binary} reports pairwise agreement under three label projections. Under the binary \textsc{yes} vs.\ non-\textsc{yes} projection reported in \cref{sec:governance}, human--labeler agreement is comparable to agreement between annotators. In a leave-one-out panel against the plurality of the other three, the labeler ranked second of four, with CIs overlapping substantially. Of the 18 labeler-vs-plurality disagreements, 12 (67\%) follow one pattern: a human marks \textsc{n/a} (regulation does not apply) while the labeler marks \textsc{no} (obligation applies and is unmet). This trigger-vs-judgment split motivated round~2.

\begin{table}[h]
\centering
\caption{Round-1 pairwise agreement on the 75-item governance subset under three label projections. The \textsc{yes} vs.\ $\neg$\textsc{yes} projection collapses the \textsc{n/a}/\textsc{no} distinction and is the one reported in \cref{sec:governance}, so only it shows 95\% CIs (Wilson on agreement, bootstrap on $\kappa$).}
\label{tab:human-iaa-binary}
\begin{tabular}{lccc}
\toprule
Pair & Ternary agreement ($\kappa$) & \textsc{yes} vs.\ $\neg$\textsc{yes} ($\kappa$) & \textsc{no} vs.\ $\neg$\textsc{no} ($\kappa$) \\
\midrule
A1 -- A2      & .67 (.51) & .92 [.84,.96] (.82 [.67,.94]) & .67 (.28) \\
A1 -- A3      & .73 (.60) & .88 [.79,.94] (.73 [.55,.88]) & .77 (.41) \\
A2 -- A3      & .65 (.49) & .88 [.79,.94] (.72 [.54,.88]) & .71 (.38) \\
A1 -- Labeler & .81 (.72) & .92 [.84,.96] (.82 [.69,.94]) & .81 (.54) \\
A2 -- Labeler & .65 (.48) & .87 [.77,.93] (.69 [.51,.86]) & .67 (.30) \\
A3 -- Labeler & .69 (.54) & .88 [.79,.94] (.73 [.55,.88]) & .72 (.35) \\
\bottomrule
\end{tabular}
\end{table}

\paragraph{Round 2: rubric-controlled re-annotation.}
Of the 38 round-1 contested items, the 35 \emph{without} a unanimous round-1 human-vs-labeler disagreement were independently re-labeled by all three annotators, with regulatory trigger and obligation conditions inline, the same scaffolding the labeler prompt encoded.

\paragraph{Post-rubric agreement on the contested subset.}
With the rubric, the annotators largely agree with one another on the 35 contested items (ternary $\kappa$ of $0.74$, $0.73$, and $0.91$ for the three pairs) and moderately with the labeler ($0.57$--$0.61$).

\paragraph{Post-rubric candidate labeler errors.}
5/75 items have unanimous post-rubric human disagreement with the labeler ($6.7\%$, 95\% Wilson CI $[2.9, 14.7]$); four more have post-rubric 2-of-3 plurality disagreement. Counting these too gives a candidate-error rate of $12.0\%$ (95\% Wilson CI $[6.4, 21.3]$).

\subsection{CriteriaBank: Human Validation}
\label{app:human-validation-aspect}

\paragraph{Setup.}
Human validation on CriteriaBank replicates the protocol of \cref{app:human-validation} on a 75-item sample drawn from CriteriaBank's broad-coverage subset, stratified across the three ternary streams (Stream~A: 25, Stream~B: 26, Stream~C: 24). The same three co-author annotators labeled each item in $\{\textsc{yes}, \textsc{no}, \textsc{n/a}\}$ under the same blind protocol. We compare their labels with two model label sources: the generator's candidate labels (``teacher'' below) and the final judge-panel labels (\cref{app:labeling}). Unlike governance, CriteriaBank teacher prompts did not encode domain-specific trigger conditions. The criteria are aspect-style yes/no questions about target text content (e.g., ``Did the agent call a tool?'').

\begin{table}[h]
\centering
\caption{Pairwise agreement on the 75-item CriteriaBank subset under the same three label projections as \cref{tab:human-iaa-binary}. CIs (Wilson 95\% on agreement, bootstrap 95\% on $\kappa$, $B{=}2000$) are shown for the \textsc{yes} vs.\ $\neg$\textsc{yes} projection only. A2 missing on 1 item drops $n$ for affected pairs.}
\label{tab:aspect-iaa-binary}
\begin{tabular}{lccc}
\toprule
Pair & Ternary agreement ($\kappa$) & \textsc{yes} vs.\ $\neg$\textsc{yes} ($\kappa$) & \textsc{no} vs.\ $\neg$\textsc{no} ($\kappa$) \\
\midrule
A1 -- A2      & .81 (.71) & .89 [.80,.94] (.77 [.60,.91]) & .85 (.70) \\
A1 -- A3      & .81 (.70) & .87 [.77,.93] (.71 [.54,.86]) & .87 (.73) \\
A2 -- A3      & .76 (.62) & .86 [.77,.92] (.72 [.55,.86]) & .80 (.59) \\
A1 -- Teacher & .73 (.60) & .93 [.85,.97] (.85 [.70,.97]) & .76 (.52) \\
A2 -- Teacher & .70 (.55) & .91 [.82,.95] (.80 [.64,.93]) & .78 (.53) \\
A3 -- Teacher & .71 (.56) & .88 [.79,.94] (.74 [.58,.89]) & .79 (.56) \\
\bottomrule
\end{tabular}
\end{table}

\paragraph{Aggregate agreement.}
Under the \textsc{yes} vs.\ non-\textsc{yes} projection, human--teacher agreement is comparable to agreement between annotators and to the governance subset (\cref{tab:aspect-iaa-binary}). Under ternary labels, the teacher agrees less often than annotators agree with each other, driven by its use of \textsc{n/a}. In a leave-one-out panel, the teacher ranks last of four ($75.3\%$ [64, 84] vs.\ $81.9$--$87.3\%$ for the annotators), the reverse of the governance ordering, though all CIs overlap.

\paragraph{Disagreement direction and confirmed errors.}
Of the 18 items where the teacher disagrees with the human plurality, 14 (78\%) involve teacher \textsc{n/a} opposite a committed human verdict (11 \textsc{no} vs.\ \textsc{n/a}, 3 \textsc{yes} vs.\ \textsc{n/a}). This is the opposite of governance, where the labeler more often \emph{commits} where humans \emph{abstain}. Both panels concentrate disagreement on the same \textsc{n/a}-vs-committed axis. Of the 18, 9 are unanimous human-vs-teacher; A1 re-examined each post-hoc and judged the teacher label wrong on all 9 (none borderline), giving a confirmed-error rate of $9/75=12.0\%$ (95\% Wilson CI $[6.4, 21.3]$). The 9 plurality-only items are unadjudicated, bringing the candidate-error upper bound to $18/75=24.0\%$. 

\paragraph{Judge-panel labels.}
We score the final judge-panel labels (\cref{app:labeling}) against the same blind human panel. Of the 75 audited items, 71 could be matched to final-corpus records (counterfactual-augmentation pairs share a criterion and target and were matched on context overlap; 4 items remained ambiguous and are excluded), and 2 of these 71 have a three-way annotator split and no majority label. On the remaining 69, agreement with the human majority label is $88.4\%$ for the panel labels versus $73.9\%$ for the teacher labels (paired exact McNemar $p=0.013$), and $95.8\%$ versus $81.2\%$ on the $n{=}48$ subset where all three annotators agree. Of the 9 confirmed teacher errors, the panel corrects 8. Such errors carry over to the model: on the same 9 items, a VFM-4B pretrained on the teacher labels predicts the teacher label in all 9 cases, while a 4B model pretrained on the panel labels (without KL regularization) follows the human majority in 7. Most of the gain comes from items where the property is absent, which the teacher often labeled \textsc{n/a} instead of \textsc{no}.

\subsection{INSURE-Dial: Real-Data Validation Details}
\label{app:insure-dial}

\paragraph{Evaluation set and protocol.}
INSURE-Dial Task 2~\citep{kulkarni2026insuredial} evaluates each call with a single structured query that fills a fixed schema of per-phase compliance fields, such as ``Did the assistant ask the user to confirm coverage during the coverage-check phase?'' We instead derive 9 atomic criterion templates from the dataset's annotations, giving 6 or 9 templates per call (the three drug-specific templates repeat for each drug discussed) and 417 records in total across the 50 test calls (47.5\% YES, 18.9\% NO, 33.6\% \textsc{n/a}). Seven templates correspond to schema fields and yield 317 records. The other two, greeting (``Did the assistant provide a greeting?'') and reference-number elicitation (``Did the assistant ask the user for a reference number?''), have no schema field, so the source paper's protocol cannot score them. For the GPT-5.4 paper-protocol baseline, we use the source paper's code,\footnote{\url{https://github.com/interactly-aimon/insuredial-dataframer-eacl-2026}} prompt, and schema.

\paragraph{Training composition.}
We finetune on calls from INSURE-Dial's synthetic training split, rebalanced to about 2K records (785 YES, 430 NO, and 785 \textsc{n/a}), keeping every available NO record. %

\paragraph{Checkpoint selection.}
For the headline results (\cref{tab:insure-dial}), each model is trained for a fixed 20 epochs (3 seeds), and we select epochs by 5-fold cross-validation over the 50 real calls, split by call. For each fold, we select the epoch by accuracy on the other 40 calls and use it to predict the 10 held-out calls. Selecting on macro-F1 instead changes no mean accuracy by more than 0.4 points. We select on real calls because macro-F1 on the synthetic dev split saturates after about 3 epochs, while accuracy on real calls keeps improving.

\paragraph{Frontier baselines.}
We evaluate GPT-5.4 (low reasoning) in two ways that cover all 417 records. The per-criterion baseline asks about one criterion at a time. The extended-schema baseline adds greeting and reference-number fields to the source paper's schema and fills the whole schema in one query per call.

\paragraph{Unfinetuned baselines.}
Without finetuning, the pretrained VFM almost never predicts \textsc{n/a} on INSURE-Dial, so all 140 \textsc{n/a} records count as errors. The Qwen3-4B judge receives the same prompt as the per-criterion GPT-5.4 baseline and answers directly. It too rarely answers \textsc{n/a} and labels most \textsc{n/a} records \textsc{no}, whereas GPT-5.4 with the same prompt labels most of them correctly.

\paragraph{Violation detection.}
Aggregate accuracy hides a failure mode: the finetuned 4B models label $19$--$26\%$ of gold violations as satisfied. Confidence does not identify these errors. Most of the VFM-4B's false passes have max-softmax of at least $0.9$, and 4 of the 37 violating calls are missed by all three seeds, so seed disagreement cannot flag them. Reviewing the $30\%$ of calls with the most uncertain verdicts raises the share of violating calls with at least one flagged violation only from $84.7\%$ to $87.4\%$, and full recall requires reviewing every call in which the VFM flags no violation ($34\%$ of calls). Performance on INSURE-Dial is therefore not sufficient for standalone enforcement.

\subsection{DynaBench: Details}
\label{app:dynabench}

\paragraph{Rule decomposition pipeline.}
DynaBench rules are natural-language assertions, sometimes compound (e.g., ``When the user asks $X$, do $Y$ and otherwise refuse''). Before training and evaluation, GPT-5.4 translates each rule, from its text alone, into one or more atomic questions phrased so that YES means a violation, with one question per branch of a compound rule. For example, ``Never use emojis'' becomes ``Did the agent use emojis?'' The translator never sees traces or labels, and its translations average $1.52$ questions per rule. For training, each question inherits its rule's label.

\paragraph{Clause-splitting ablation.}
To measure how much the VFM relies on decomposition, we keep one DynaBench-finetuned 4B checkpoint fixed (from an earlier configuration without KL regularization) and replace each rule's atomic questions with a single whole-rule question. Trace-level FAIL-F1 drops from $0.850$ to $0.727$, partly because several questions give more chances to detect a violation. We release the translation prompts and the translated criteria.

\paragraph{Aggregation and localization.}
A trace is predicted FAIL if any of its questions has a violation probability, renormalized over YES and NO, above the dev-selected threshold. Every FAIL trace in DynaBench violates exactly one rule. To localize it, we score each rule by its highest question-level violation probability and check whether the violated rule ranks first (top-1) or among the top three (top-3). Top-1 localization is $97.6\%$ for VFM-4B, and top-3 is $100\%$ for every model and seed, so the cascade's top three always include the violated rule. Replacing them with three random rules drops the cascade to $0.675$ F1. The cascade passes the raw text of these rules to the Gemma judge, the input format under which that judge is strongest (\cref{tab:dynabench}). Because DynaBench rules always apply to their traces, renormalization costs nothing here. In \textsc{n/a}-heavy settings it can rank an inapplicable criterion above a true violation, as the next paragraph shows.

\paragraph{Multi-violation localization.}
Governance and INSURE-Dial score each trace against many criteria, and most traces with a violation there violate two or more (35 of 37 INSURE-Dial calls and 38 of 42 governance interactions). On these traces, we rank each trace's criteria by violation probability. For VFM-4B, the top three almost always include a true violation (\cref{tab:multiviol}), so the cascade's shortlist still works when several criteria fail. Top-1 is high on governance but lower on INSURE-Dial, mostly because criteria the VFM correctly predicts as \textsc{n/a} can rank first after renormalization. Ranking only criteria with a gold YES or NO label raises INSURE-Dial top-1 to $96.2\%$.

\begin{table}[h]
\centering
\small
\caption{\textbf{Multi-violation localization} (VFM-4B, 3-seed mean) on the $k{\ge}2$ subset (traces violating $\ge2$ criteria) of two per-criterion benchmarks. Top-1/top-3 ask whether the highest / top-3 ranked criteria include a true violation, recall@$k$ is the fraction of the $k$ violations in the top-$k$, all-in-top-$k$ requires all of them.}
\label{tab:multiviol}
\begin{tabular}{@{}lccccc@{}}
\toprule
Benchmark & $k{\ge}2$ traces & top-1 & top-3 & recall@$k$ & all-in-top-$k$ \\
\midrule
INSURE-Dial & $35$ & $0.676$ & $0.933$ & $0.666$ & $0.505$ \\
Governance  & $38$ & $0.956$ & $1.000$ & $0.824$ & $0.342$ \\
\bottomrule
\end{tabular}
\end{table}

\paragraph{Cascade lift by criteria count.}
\Cref{tab:dynabench-bins} stratifies F1 by atomic-criteria count $|C|$ per trace. For traces with few rules ($|C|{\le}10$), the cascade is marginally below Gemma alone, and its lift grows with $|C|$.

\begin{table}[h]
\centering
\small
\caption{F1 stratified by $|C|$ (atomic criteria per trace), 3-seed means (the KL-VFM cascade of \cref{tab:dynabench} and the 3-seed Gemma-only baseline on the same 543 traces). ``\% FAIL'' is the ground-truth fraction.}
\label{tab:dynabench-bins}
\begin{tabular}{lrrrrr}
\toprule
$|C|$ stratum & traces & \% FAIL & Gemma-only CoT & Cascade top-3 CoT & $\Delta$ \\
\midrule
$\le 10$ & 220 & 50.0\% & $0.944$ & $0.930$ & $-1.4$\,pp \\
$11$--$20$ & 149 & 32.2\% & $0.898$ & $0.941$ & $+4.4$\,pp \\
$21$--$40$ & 112 & 42.0\% & $0.898$ & $\mathbf{0.952}$ & $\mathbf{+5.4}$\,pp \\
$>40^{\dag}$ & 62 & 100.0\% & $0.919$ & $0.975$ & $+5.7$\,pp \\
\midrule
all & 543 & 49.2\% & $0.922$ & $0.946$ & $+2.4$\,pp \\
\bottomrule
\end{tabular}
\\[2pt]
{\footnotesize $^{\dag}$The $|C|{>}40$ stratum is $100\%$ FAIL by ground truth (all $62$ traces are failures); we report it for completeness.}
\end{table}

\paragraph{Selective escalation.}
The cascade forwards the top three rules for every trace. Alternatively, the VFM can decide the traces it is most confident about and forward only the least-confident fraction $e$ to the Gemma judge. \Cref{tab:dynabench-selective} reports FAIL-F1 on the test set as a function of $e$, with the VFM's threshold also tuned on the test set (so the $e{=}0$ point is slightly above the $0.885$ of \cref{tab:dynabench}). Escalating about a fifth of traces matches the all-rules Gemma baseline, and full escalation recovers the cascade.

\begin{table}[h]
\centering
\small
\caption{\textbf{Selective escalation on the DynaBench test set} (KL-VFM-4B ranker, 3 seeds). Escalating the least-confident fraction $e$ of traces to the Gemma-4-26B judge and deciding the rest with the VFM alone. $e{=}0$ is the VFM standalone, $e{=}1$ the full top-3 cascade; $e$ was not selected on a separate development set.}
\label{tab:dynabench-selective}
\begin{tabular}{lrr}
\toprule
Escalated fraction $e$ & Traces to judge & FAIL-F1 \\
\midrule
$0.00$ (VFM alone) & $0$   & $0.893 \pm 0.006$ \\
$0.15$             & $81$  & $0.917 \pm 0.005$ \\
$0.25$             & $136$ & $0.928 \pm 0.002$ \\
$0.50$             & $272$ & $0.937 \pm 0.004$ \\
$1.00$ (full cascade) & $543$ & $0.946 \pm 0.005$ \\
\bottomrule
\end{tabular}
\end{table}

\paragraph{Holistic-judge rule-input ablation.}
\Cref{tab:dynabench} also gives the Gemma judge either the raw rules, as in DynaGuard's protocol, or the VFM's atomic questions. The atomic questions lower its FAIL-F1 by 12 points, almost entirely through recall, because with question-form rules Gemma becomes reluctant to return FAIL. Decomposition is therefore not a general improvement for any judge, but a representation suited to the VFM's per-criterion head.

\paragraph{External baseline notes.}
We cite DynaGuard's reported $0.720$ directly. Our numbers use the standard 543-trace test split of \texttt{tomg-group-umd/DynaBench}, the default in DynaGuard's evaluation script. GPT-5.4, used for translation, saw only rule text and so cannot leak compliance labels. The Gemma 4 baselines, released in 2026, may have seen DynaBench during training.

\paragraph{Per-trace latency vs.\ DynaGuard.}
We also time single traces on an H100 with HuggingFace Transformers, using 5 warm-up and 20 measured test traces. VFM-4B encodes each trace once and evaluates its criteria against the cached prefix. DynaGuard-4B runs in its fast mode, which returns a short verdict without chain-of-thought. VFM-4B returns all of its per-criterion verdicts in a median of 79\,ms, about $1.6{\times}$ faster per trace than DynaGuard-4B returns its single PASS/FAIL verdict. DynaGuard's published $11.6$\,ms per example uses batched vLLM serving and is not comparable.

\subsection{Violation-Centric Metrics}
\label{app:violation-metrics}

\Cref{tab:violation-metrics} reports violation precision, recall, and F1 alongside the corresponding frontier-judge results. For Governance and INSURE-Dial, a violation is a criterion-level \textsc{no}. For DynaBench, it is a trace-level FAIL. The final column is the fraction of gold violations that the VFM predicts as passes (gold \textsc{no}$\to$\textsc{yes} for the criterion-level tasks, and gold FAIL$\to$PASS for DynaBench).

\begin{table}[t]
\centering
\footnotesize
\setlength{\tabcolsep}{3pt}
\caption{\textbf{Violation-centric results across governance benchmarks.} P/R/F1 are the precision, recall, and F1 of detecting a violation, in percent. VFM-4B results are means over three seeds. GPT-5.4 rows are single deterministic runs (on INSURE-Dial, with the extended schema), and the Gemma-4-26B row, given the raw DynaBench rules, is the mean over three deterministic runs.}
\label{tab:violation-metrics}
\begin{tabular*}{\textwidth}{@{\extracolsep{\fill}}lclcc@{}}
\toprule
Benchmark & VFM-4B P\,/\,R\,/\,F1 & Frontier judge & Judge P\,/\,R\,/\,F1 & VFM pass error \\
\midrule
Governance (in-distribution) & 83.8\,/\,87.4\,/\,85.5 & GPT-5.4 & 82.8\,/\,77.9\,/\,80.3 & 5.6\% \\
Governance (held-out regulations) & 76.7\,/\,73.0\,/\,74.6 & GPT-5.4 & 76.8\,/\,75.5\,/\,76.1 & 11.1\% \\
INSURE-Dial (real calls) & 82.4\,/\,75.1\,/\,78.6 & GPT-5.4 & 60.2\,/\,86.1\,/\,70.8 & 19.4\% \\
DynaBench (traces) & 83.1\,/\,94.9\,/\,88.5 & Gemma-4-26B & 96.0\,/\,88.8\,/\,92.2 & 5.1\% \\
\bottomrule
\end{tabular*}
\end{table}

Because the tree mask makes each criterion's verdict independent of the others (\cref{sec:architecture}), adding criteria cannot hide an existing violation, only add chances for a false alarm. On DynaBench, FAIL recall stays at or above $91.5\%$ in every criterion-count bin, with no degradation as the number of criteria grows.

\subsection{Compute}
\label{app:compute}

All training, evaluation, and latency experiments ran on NVIDIA H100 80\,GB GPUs (bf16). CriteriaBank pretraining used $4{\times}$H100 at both model sizes. CriteriaBank pretraining was 1 epoch on the full $393{,}531$-record corpus ($3{,}075$ steps at global batch size $128$). Estimated compute is on the order of ${\sim}48$--$96$ GPU-hours per VFM-4B run and ${\sim}12$--$24$ GPU-hours per VFM-0.6B run. Each downstream finetuning run takes 30--90 minutes on $4{\times}$H100 for the 4B models (with shorter runs at 0.6B), and aggregate downstream finetuning across the eight benchmarks and three seeds each is on the order of ${\sim}100$ GPU-hours. External LLM-judge baselines were evaluated via API or vLLM batched serving, and aggregate API spend across all benchmarks was on the order of \$$100$--\$$300$. Including preliminary experiments, ablations not reported in the paper, and abandoned design iterations, the full research project exceeded the reported figures by an estimated factor of $2$--$3{\times}$.

\section{Decomposition Analysis: How Many Rules Govern a Real Deployment?}
\label{app:decomposition_analysis}

This appendix supports the claim in \cref{sec:background} that real deployments are governed by hundreds of distinct rules. We work through two examples: the regulatory surface of the New York City \textsc{MyCity} chatbot (\cref{app:mycity-decomposition}), and one section of the Claude Opus~4.7 system prompt (\cref{app:claude-decomposition}).

\subsection{The \textsc{MyCity} Chatbot: A Retrospective Rule-Count Estimate}
\label{app:mycity-decomposition}

Investigative reporting documented incorrect legal advice from the NYC \textsc{MyCity} chatbot across multiple domains including housing discrimination, tenant lockouts, rent stabilization, cash-payment rules, tip retention, scheduling notices, and funeral-service disclosures~\citep{themarkup2024mycity,themarkup2024mycity2}. Five representative examples are:

\begin{decompbox}
1. Does the response state or imply that an employer or its agent may demand, accept, or retain any part of an employee’s gratuities, contrary to N.Y. Labor Law 196-d?

2. Does the response state or imply that covered landlords, owners, or brokers may refuse housing because a prospective or current tenant uses a housing voucher or other lawful source of income, contrary to NYC Admin. Code 8-102 and 8-107(5)?

3. Does the response state or imply that a covered food store or retail establishment may refuse cash payments from consumers, contrary to NYC Admin. Code 20-840, enacted by Local Law 34 of 2020?

4. Does the response state or imply that rent on rent-stabilized units is unrestricted, contrary to the Rent Stabilization Law’s legal regulated rent provisions, including NYC Admin. Code 26-512?

5. Does the response state or imply that a landlord or other person may lock out a protected residential occupant without a warrant of eviction, court order, or governmental vacate order, contrary to RPAPL 768 and NYC Admin. Code 26-521?
\end{decompbox}

These criteria were extracted from news reports~\citep{themarkup2024mycity,themarkup2024mycity2} and matched to specific legal provisions with the assistance of GPT-5.5.

These domains are only a slice of the regulatory surface a small-business advisory chatbot must respect. To estimate the broader scope, we used an LLM (Gemini 3 Pro) to enumerate candidate legal and governance frameworks for the deployment, running multiple passes across regulatory categories and jurisdictional layers (NYC, New York State, and federal) and deduplicating across passes. Single-pass enumeration yielded $21$ frameworks; multi-pass deduplication surfaced $141$. We then drew a random sample of $30$ frameworks and asked the model to enumerate provisions within each, finding an average of roughly $15$ provisions per framework; extrapolating to all $141$ yields roughly $2{,}100$ atomic provisions in total. A targeted recovery check confirmed that the full set of documented violation areas extracted from the reporting was surfaced; single-pass enumeration missed five of those documented areas.

A complete governance specification for a \textsc{MyCity}-style deployment thus plausibly involves on the order of $2{,}000$ atomic provisions, most inapplicable to any given user query.

\subsection{One Section of Claude's System Prompt}
\label{app:claude-decomposition}

Anthropic publishes the core system prompts used in the Claude web and mobile interfaces, while noting that these prompts do not apply to the Claude API~\citep{anthropic2026claudeprompt}. We illustrate an atomic-rule decomposition on a portion of the ``User Wellbeing'' section of the Claude Opus~4.7 system prompt below.

\textbf{Source excerpt.}
\begin{promptbox}
Claude uses accurate medical or psychological information or terminology where relevant.

Claude cares about people's wellbeing and avoids encouraging or facilitating self-destructive behaviors such as addiction, self-harm, disordered or unhealthy approaches to eating or exercise, or highly negative self-talk or self-criticism, and avoids creating content that would support or reinforce self-destructive behavior even if the person requests this. Claude should not suggest techniques that use physical discomfort, pain, or sensory shock as coping strategies for self-harm (e.g. holding ice cubes, snapping rubber bands, cold water exposure), as these reinforce self-destructive behaviors. [...] 

In ambiguous cases, Claude tries to ensure the person is happy and is approaching things in a healthy way.

[...]
\end{promptbox}

\textbf{Decomposition.}
\begin{decompbox}
1. If the response includes medical or psychological information or terminology, is it accurate?

2. Does the response encourage self-destructive behavior such as addiction, self-harm, disordered eating or exercise, or highly negative self-talk or self-criticism?

3. Does the response facilitate self-destructive behavior, e.g., by providing instructions, methods, or other content?

4. Does the response produce content that supports or reinforces self-destructive behavior?

5. Does the response suggest physical-discomfort or sensory-shock techniques (e.g., holding ice cubes, snapping rubber bands, cold water exposure) as coping strategies for self-harm?

6. If the user's wellbeing is ambiguous, does the response make some attempt to ensure the user is happy and is approaching things in a healthy way?
\end{decompbox}

This decomposition was proposed by an LLM and manually edited by a human with legal training. 

This roughly $115$-word fragment yields six criteria, many of which might be further decomposed, a density of about one criterion per twenty words. If this density were representative of the full system prompt, a several-thousand-word prompt would imply hundreds of atomic behavioral rules. The Claude Code or Cursor harness adds further tool-use protocols, project conventions, memory entries, and user-specific instructions on top, expanding the per-session deployment surface to $>$15{,}000 words.

\end{document}